%% file: root.tex
\documentclass[lettersize,journal]{IEEEtran}
\usepackage{amsmath,amsfonts,amssymb}

\usepackage{algorithmic}
\usepackage{algorithm}
\usepackage{array}
\usepackage[caption=false,font=normalsize,labelfont=sf,textfont=sf]{subfig}
\usepackage{textcomp}
\usepackage{stfloats}
\usepackage{url}
\usepackage{verbatim}
\usepackage{graphicx}
\usepackage{cite}
\usepackage{xcolor}
\usepackage{threeparttable}
\usepackage{hyperref}
\usepackage{arydshln}
\input{user_commands}

\begin{document}

\title{OKVIS2-X: Open Keyframe-based Visual-Inertial SLAM Configurable with Dense Depth or LiDAR, and GNSS}

\author{Simon Boche$^{*}$, Jaehyung Jung$^{*}$, Sebastián Barbas Laina$^{*}$, and Stefan Leutenegger
    \thanks{This work was supported by the EU Horizon Europe program under grant agreement 101070405 (DigiForest) and 101120732 (AUTOASSESS).}
    \thanks{Simon Boche, Jaehyung Jung and Sebastián Barbas Laina are with the Mobile Robotics Lab,  School of Computation, Information and Technology (CIT) at the Technical University of Munich (TUM), 80333 Munich, Germany,  and with the Munich Institute of Robotics and Machine Intelligence (MIRMI). (e-mail: simon.boche@tum.de; jaehyung.jung@tum.de; sebastian.barbas@tum.de) \\
    Stefan Leutenegger is with the Mobile Robotics Lab, ETH Zurich, 8092 Zurich, Switzerland. (e-mail: lestefan@ethz.ch). }
    \thanks{$^{*}$ Equal contribution.}
}



\begin{minipage}{\textwidth}
This paper has been accepted for publication in the Special Issue on Visual SLAM of \textit{IEEE Transactions on Robotics (T-RO).} The final version will be available via IEEE Xplore.

\vspace{2em}

\centerline{DOI: 10.1109/TRO.2025.3619051} 

\vspace{2em}

\copyright 2025 IEEE. Personal use of this material is permitted. Permission from IEEE must be obtained for all other uses, in any current or future media, including reprinting/republishing this material for advertising or promotional purposes, creating new collective works, for resale or redistribution to servers or lists, or reuse of any copyrighted component of this work in other works.
\end{minipage}

\markboth{IEEE Transactions on Robotics (T-RO) - Special Issue: Visual SLAM. Preprint Version. Accepted August 2025}
{Boche \MakeLowercase{\textit{et al.}}: OKVIS2-X: Open Keyframe-based Visual-Inertial SLAM Configurable with Dense Depth or LiDAR, and GNSS} 

\maketitle

\begin{abstract}

To empower mobile robots with usable maps as well as highest state estimation accuracy and robustness, we present OKVIS2-X: a state-of-the-art multi-sensor Simultaneous Localization and Mapping (SLAM) system building dense volumetric occupancy maps, while scalable to large environments and operating in realtime.
Our unified SLAM framework seamlessly integrates different sensor modalities: visual, inertial, measured or learned depth, LiDAR and Global Navigation Satellite System (GNSS) measurements. 
Unlike most state-of-the-art SLAM systems, we advocate using dense volumetric map representations when leveraging depth or range-sensing capabilities. We employ an efficient submapping strategy that allows our system to scale to large environments, showcased in sequences of up to 9 kilometers. 
OKVIS2-X enhances its accuracy and robustness by tightly-coupling the estimator and submaps through map alignment factors. Our system provides globally consistent maps, directly usable for autonomous navigation.
To further improve the accuracy of OKVIS2-X, we also incorporate the option of performing online calibration of camera extrinsics.
Our system achieves the highest trajectory accuracy in EuRoC against state-of-the-art alternatives, outperforms all competitors in the Hilti22 VI-only benchmark, while also proving competitive in the LiDAR version, and showcases state of the art accuracy in the diverse and large-scale sequences from the VBR dataset.
Code available at: \href{https://github.com/ethz-mrl/OKVIS2-X}{https://github.com/ethz-mrl/OKVIS2-X}.
\end{abstract}

\begin{figure}[!t]
    \centering
    \includegraphics[width=\linewidth]{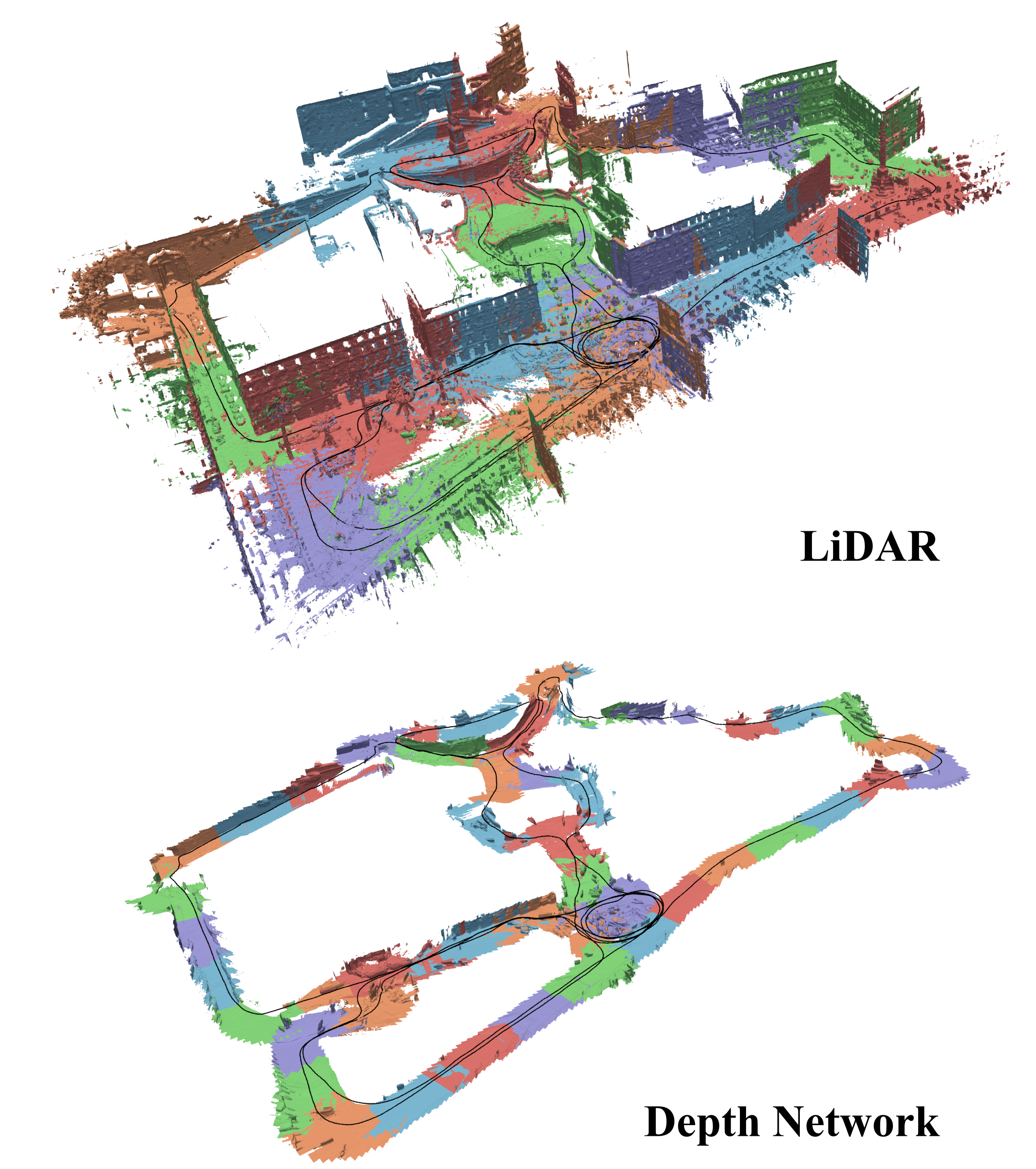}
    \caption{3D reconstruction from a run of OKVIS2-X on the \texttt{Spagna} sequence of the VBR dataset~\cite{VBR}. Reconstruction with a LiDAR sensor (top) or with a depth network (bottom) to showcase the versatility of the presented system to different sensor modalities. The estimated trajectory is visualized in black. Furthermore, different colors per submap are used.}
    \label{fig:top-figure}
\end{figure}

\begin{IEEEkeywords}
SLAM; Mapping; Localization; Sensor Fusion
\end{IEEEkeywords}

\input{chapters/01_introduction}
\input{chapters/02_related-work}
\input{chapters/03_preliminaries}
\input{chapters/04_OccMapping}
\input{chapters/05_SLAM}
\input{chapters/06_evaluation}
\input{chapters/07_conclusion}

\input{chapters/99_appdx}


\bibliographystyle{IEEEtran}
\bibliography{references}

\begin{IEEEbiography}[{\includegraphics[width=1in,height=1.25in,clip,keepaspectratio]{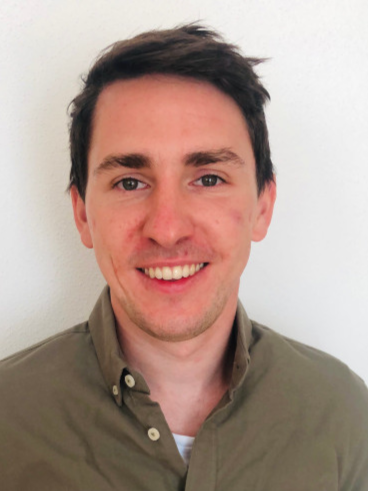}}]{Simon Boche}
is currently pursuing a Ph.D. degree at the Mobile Robotics Lab (MRL) at the Technical University of Munich (TUM), Germany, under the supervision of Prof. Dr. Stefan Leutenegger.
He received the M.Sc. degree in Robotics, Cognition, Intelligence and the B.Sc. degree in Engineering Science, both from TUM.
His research interests include multi-sensor fusion, localization and mapping, and autonomous navigation for mobile robots.
\end{IEEEbiography}
\begin{IEEEbiography}[{\includegraphics[width=1in,height=1.25in,clip,keepaspectratio]{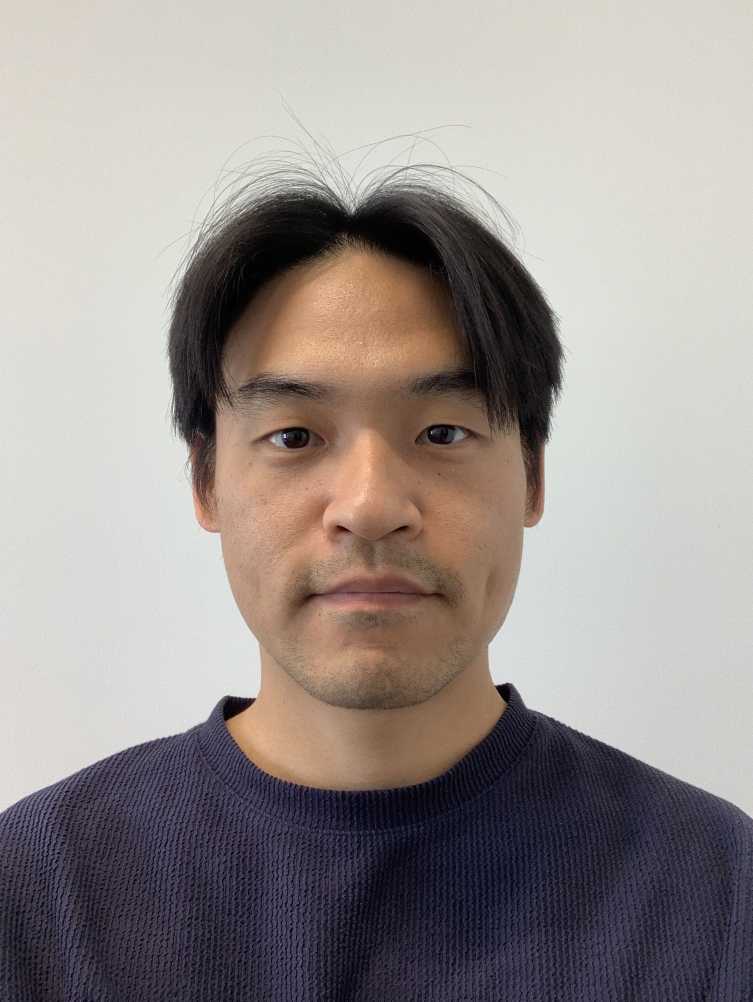}}]{Jaehyung Jung}
is currently a Postdoctoral Researcher at the Mobile Robotics Lab (MRL) at the Technical University of Munich (TUM). 
He received the Ph.D. and M.Sc. degrees in Aerospace Engineering from Seoul National University, and the B.Sc. degree in Aerospace Engineering from Pusan National University.  
His research interests include state estimation and robot perception.
\end{IEEEbiography}
\begin{IEEEbiography}[{\includegraphics[width=1in,height=1.25in,clip,keepaspectratio]{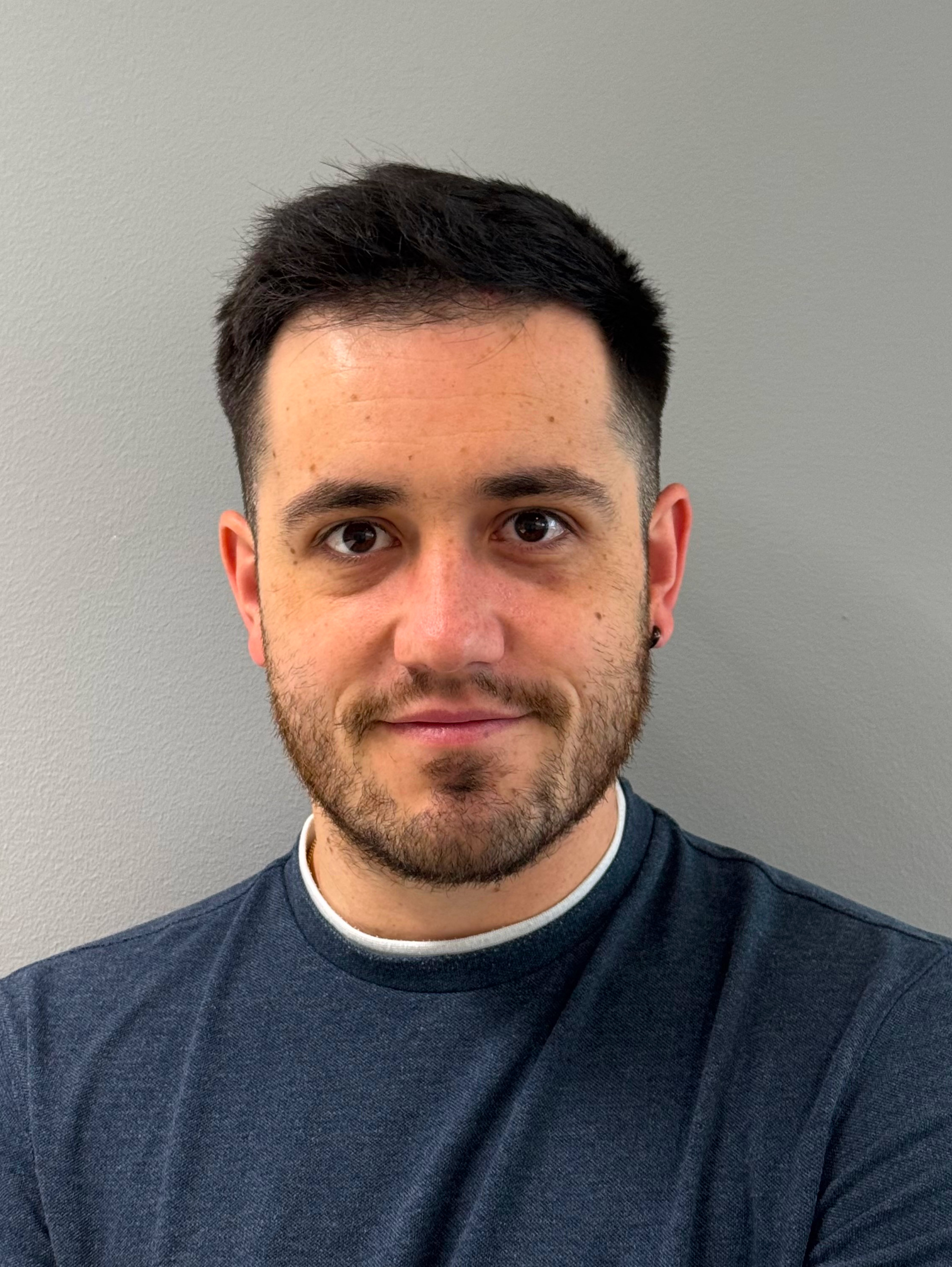}}]{Sebastián Barbas Laina}
is currently pursuing a Ph.D. degree at the Mobile Robotics Lab (MRL) at the Technical University of Munich (TUM), under the supervision of Prof. Dr. Stefan Leutenegger. 
He received the M.Sc. degree in Systems, Control and Robotics from KTH Royal Institute of Technology, and the B.Sc. degree in Industrial Engineering with a focus on Electrical Engineering from Universidad Politécnica de Madrid (UPM).
His research interests include localization and mapping, multi-sensor fusion, and autonomous navigation for mobile robots.
\end{IEEEbiography}
\begin{IEEEbiography}[{\includegraphics[width=1in,height=1.25in,clip,keepaspectratio]{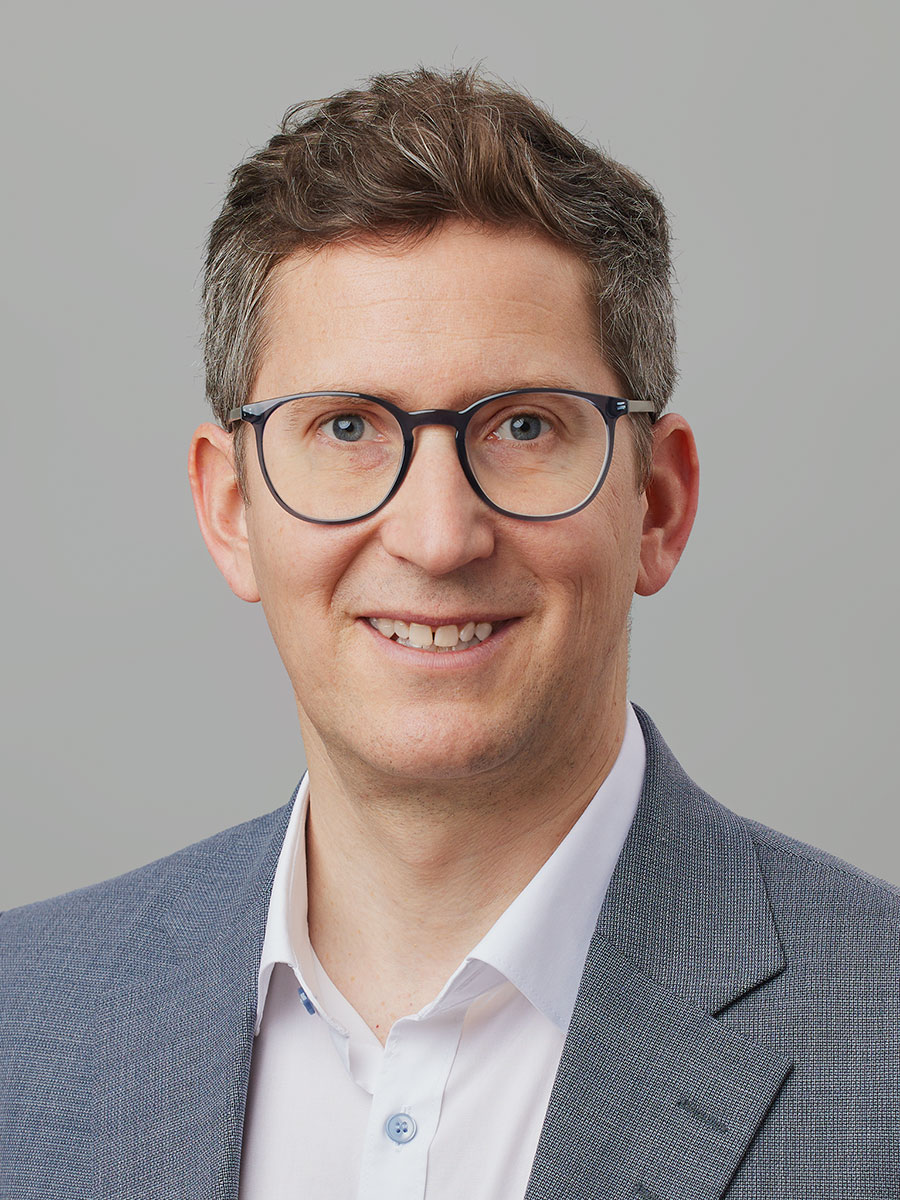}}]{Stefan Leutenegger} has been an Associate Professor of Mobile Robotics in the Department of Mechanical and Process Engineering at ETH Zurich since early 2025. Prior to that, in 2021, he joined the Technical University of Munich (TUM) as a Tenure-Track Assistant Professor. Between 2014 and 2020, he was a (Senior) Lecturer at Imperial College London, where he continues to hold an honorary Reader position.
\end{IEEEbiography}

\input{chapters/999_supplementary}


\end{document}

%% file: user_commands.tex
\usepackage{xifthen}
\usepackage{xparse}
\usepackage{leftidx}
\usepackage{bm}
\usepackage{etoolbox}
\usepackage{amsmath}
\usepackage{multirow, makecell}
\usepackage{xspace}

\newcommand{\abs}[1]{\left\vert#1\right\vert}

\newcommand{\bbm}{\begin{bmatrix}}
\newcommand{\ebm}{\end{bmatrix}}

\DeclareMathAlphabet{\mybf}{OT1}{ptm}{b}{n} 
\newcommand{\mybs}[1]{{\bm{#1}}} 
\DeclareMathAlphabet{\mybfi}{OML}{cmm}{b}{it}

\newcommand{\mbf}[1]{
\ifcat\noexpand#1\relax 
\mybs{#1}
\else
\mybf{#1}
\fi
}

\newcommand{\mbfbar}[1]{{\overline{\mbf{#1}}}}
\newcommand{\mbfhat}[1]{{\hat{\mbf{#1}}}}
\newcommand{\mbftilde}[1]{{\tilde{\mbf{#1}}}}

\newcommand{\mbfdot}[1]{{\dot {\mbf{#1}}}}

\NewDocumentCommand{\mbfidentity}{o}{\IfValueTF{#1}{\mbf{I}_{#1\hspace{\rightshift}}}{\mbf{I}}}
\NewDocumentCommand{\mbfzero}{oo}{\IfValueTF{#1}{\mbf{0}_{#1\times#2\hspace{\rightshift}}}{\mbf{0}}}

\newcommand{\cframe}[1]{{\smash{\protect\underrightarrow{\mathcal{F}}_{#1}}}}

\newcommand{\homo}[1]{{\mybfi{#1}}}
\newcommand{\mbfhbar}[1]{{\overline{\homo{#1}}}}

\newcommand{\mbfh}[1]{{\homo{#1}}}

\newcommand{\crossmx}[1]{\left[{#1}\right]^\times}

\newlength{\leftshift}
\newlength{\rightshift}
\newcommand{\mvec}[2]{\leftidx{_{#1}}{\mbf #2}{}} 

\newcommand{\pos}[2]{\leftidx{_{#1}}{ \mbf r}{_{#2\hspace{\rightshift}}}} 
\newcommand{\posbar}[2]{\leftidx{_{#1}}{\mbfbar r}{_{#2\hspace{\rightshift}}}} 
\newcommand{\poshat}[2]{\leftidx{_{#1}}{\mbfhat r}{_{#2\hspace{\rightshift}}}} 

\NewDocumentCommand{\vel}{moo}{
	\IfValueTF{#1}{\leftidx{_{#1}}}{}{\mbf v}{\IfValueTF{#2}{_{#2#3\hspace{\rightshift}}}{}}}
\NewDocumentCommand{\veltilde}{moo}{
	\IfValueTF{#1}{\leftidx{_{#1}}}{}{\mbftilde v}{\IfValueTF{#2}{_{#2#3\hspace{\rightshift}}}{}}}
\NewDocumentCommand{\velbar}{moo}{
	\IfValueTF{#1}{\leftidx{_{#1}}}{}{\mbfbar v}{\IfValueTF{#2}{_{#2#3\hspace{\rightshift}}}{}}}
\NewDocumentCommand{\velhat}{moo}{
	\IfValueTF{#1}{\leftidx{_{#1}}}{}{\mbfhat v}{\IfValueTF{#2}{_{#2#3\hspace{\rightshift}}}{}}}
\NewDocumentCommand{\veldot}{moo}{
	\IfValueTF{#1}{\leftidx{_{#1}}}{}{\mbfdot v}{\IfValueTF{#2}{_{#2#3\hspace{\rightshift}}}{}}}

\NewDocumentCommand{\acc}{moo}{
	\IfValueTF{#1}{\leftidx{_{#1}}}{}{\mbf a}{\IfValueTF{#2}{_{#2#3\hspace{\rightshift}}}{}}}
\NewDocumentCommand{\acctilde}{moo}{
	\IfValueTF{#1}{\leftidx{_{#1}}}{}{\mbftilde a}{\IfValueTF{#2}{_{#2#3\hspace{\rightshift}}}{}}}
\NewDocumentCommand{\accbar}{moo}{
	\IfValueTF{#1}{\leftidx{_{#1}}}{}{\mbfbar a}{\IfValueTF{#2}{_{#2#3\hspace{\rightshift}}}{}}}
\NewDocumentCommand{\acchat}{moo}{
	\IfValueTF{#1}{\leftidx{_{#1}}}{}{\mbfhat a}{\IfValueTF{#2}{_{#2#3\hspace{\rightshift}}}{}}}
\NewDocumentCommand{\accdot}{moo}{
	\IfValueTF{#1}{\leftidx{_{#1}}}{}{\mbfdot a}{\IfValueTF{#2}{_{#2#3\hspace{\rightshift}}}{}}}

\NewDocumentCommand{\rotvel}{moo}{
	\IfValueTF{#1}{\leftidx{_{#1}}}{}{\mbf $\omega$}{\IfValueTF{#2}{_{#2#3\hspace{\rightshift}}}{}}}
\NewDocumentCommand{\rotveltilde}{moo}{
	\IfValueTF{#1}{\leftidx{_{#1}}}{}{\mbftilde $\omega$}{\IfValueTF{#2}{_{#2#3\hspace{\rightshift}}}{}}}
\NewDocumentCommand{\rotvelbar}{moo}{
	\IfValueTF{#1}{\leftidx{_{#1}}}{}{\mbfbar $\omega$}{\IfValueTF{#2}{_{#2#3\hspace{\rightshift}}}{}}}
\NewDocumentCommand{\rotvelhat}{moo}{
	\IfValueTF{#1}{\leftidx{_{#1}}}{}{\mbfhat $\omega$}{\IfValueTF{#2}{_{#2#3\hspace{\rightshift}}}{}}}
\NewDocumentCommand{\rotveldot}{moo}{
	\IfValueTF{#1}{\leftidx{_{#1}}}{}{\mbfdot $\omega$}{\IfValueTF{#2}{_{#2#3\hspace{\rightshift}}}{}}}

\newcommand{\C}[2]{ {\mbf C}   {_{#1#2\hspace{\rightshift}} }     } 
\newcommand{\Cbar}[2]{{\mbfbar C}{_{#1#2\hspace{\rightshift}}}} 
\newcommand{\T}[2]{{\mbfh T}{_{#1#2\hspace{\rightshift}}}} 
\newcommand{\Tbar}[2]{{\mbfhbar T}{_{#1#2\hspace{\rightshift}}}} 
\newcommand{\q}[2]{{\mbf q}{_{#1#2\hspace{\rightshift}}}} 
\newcommand{\qbar}[2]{{\mbfbar q}{_{#1#2\hspace{\rightshift}}}} 

\newcommand{\Exp}[1]{\text{Exp}\left( #1 \right)}
\newcommand{\attime}[1]{^{ #1 }}



%% file: chapters/01_introduction.tex
\section{Introduction}
\label{sec:introduction}

While autonomous robots and mobile devices are becoming more ubiquitous, their ability to perceive and spatially relate to their environments remains crucial. To cope with potentially unkown environments, certain such applications demand that the system can simultaneously map the environment while also localizing in it.

\begin{figure*}[ht]
    \centering
    \includegraphics[width=\textwidth]{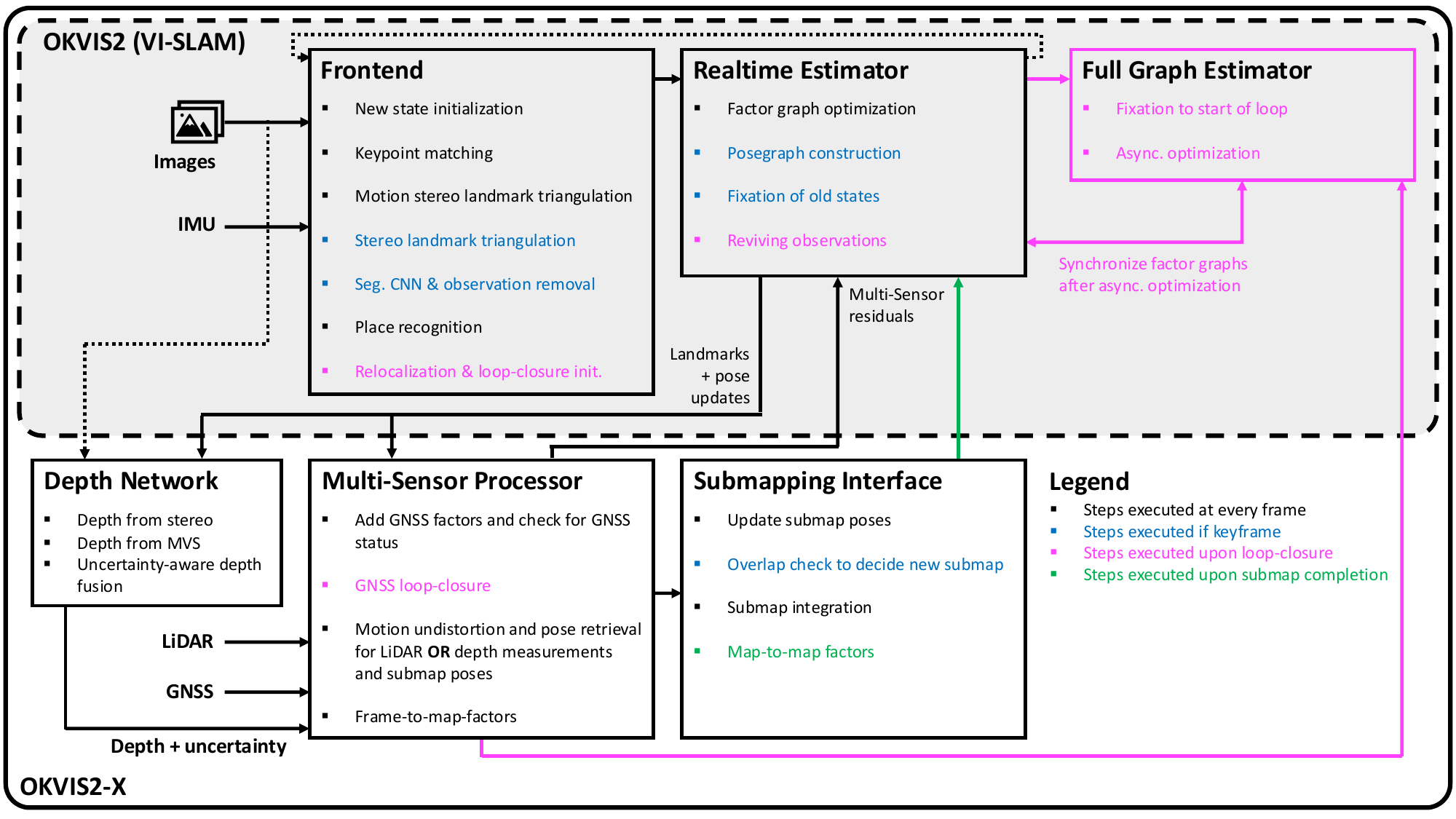}
    \caption{System architecture of our proposed multi-sensor state-estimator OKVIS2-X. Components with a grey background correspond to original elements from OKVIS2~\cite{OKVIS2} and components with a white background are the extensions for the multi-sensor setup.}
    \label{fig:system-overview}
\end{figure*}

Such Simultaneous Localization and Mapping (SLAM) systems typically require heterogeneous sensors for enhanced accuracy and robustness, with visual-inertial configurations being one of the preferred combinations of sensors due to its appealing compromise between cost and accuracy. The setup combines spatial information from the camera with proprioceptive temporal information from the inertial measurement units (IMU), whereby rendering the gravity direction observable when tightly coupled and reducing dependency on brittle visual data associations. For applications that demand a higher mapping and localization accuracy, LiDARs have emerged as a powerful option due to their unmatched depth sensing resolution and accuracy, at the expense of cost. To improve the robustness of SLAM systems, there has been a recent increase of interest in systems that can combine these three sensor modalities, whereby increasing the challenge of obtaining an accurate calibration of the extrinsics between all the different sensors, a fundamental requirement for accurate SLAM systems. Moreover, for systems that require global positioning, GNSS is the obvious choice for outdoor applications, if reliably received -- which cannot be guaranteed in all environments.

Most of the visual-inertial SLAM (VI SLAM) systems proposed in the literature \cite{campos2021orb, abate2023kimera2, geneva2020openvins, OKVIS2} only use a small subset of the camera information to build a sparse map, therefore lacking relevant geometric information required for downstream tasks.  To address this limitation, the community has proposed systems that can leverage visual depth information \cite{newcombe2011kinectfusion, tateno2017cnn, murORB2} or LiDAR information \cite{zheng2024fastlivo2, lvisam, vilens} to build denser maps that can then be reused for downstream tasks -- thereby naturally increasing memory requirements. The underlying map representation plays a crucial role not only regarding memory, but perhaps more importantly to facilitate downstream tasks. We argue that volumetric occupancy maps present an unmatched characteristic in enabling safe path planning and navigation by explicitly representing free space while the popular point clouds (or other commonly employed representations such as meshes) do not.

To address the aforementioned challenges, we propose a multi-sensor SLAM system called OKVIS2-X, a substantial extension to OKVIS2~\cite{OKVIS2}. The key idea is to leverage submap-based dense volumetric occupancy maps which are tightly-coupled to the state estimator via submap alignment factors. Examples of the 3D reconstruction obtained from our framework are presented in Fig. \ref{fig:top-figure}. Furthermore, we present a unified framework to fuse visual, inertial, LiDAR or depth from neural networks, and the global position from GNSS.
Our previous works have augmented OKVIS2 for different sensor modalities: visual-inertial-GNSS~\cite{boche2022gnss}, visual-inertial-LiDAR~\cite{boche2024tightlycoupled}, and visual-inertial-depth~\cite{jung2024uncertainty}.
But to the best of the authors' knowledge, here we present for the first time a unified and configurable multi-sensor SLAM system based on a volumetric map representation, supporting different and novel combinations of sensing modalities (e.g.\ visual-inertial-LiDAR-GNSS and visual-inertial-depth-GNSS) through a factor graph framework. In a further novel contribution, we seamlessly support online-calibration of camera extrinsics. Fig.~\ref{fig:system-overview} describes the system architecture of OKVIS2-X that is configurable with multiple sensor modalities. The resulting system demonstrates unmatched accuracy and robustness in localization and mapping, as well as scalability in building dense volumetric maps from meter to kilometer-level -- which we demonstrate in an extensive series of entirely new evaluations. The main contributions of our proposed method are:
\begin{itemize}
    \item We present a state-of-the-art multi-sensor SLAM system able to fuse visual, inertial, GNSS and dense mapping factors from either LiDAR measurements or a depth network in realtime.
    \item Our method tightly couples the state estimator with volumetric occupancy mapping, rendering it suitable for large-scale SLAM thanks to our submap representation.
    \item Our method supports online calibration of the camera-IMU extrinsics in the visual factors including relative pose error terms.
    \item We perform a thorough evaluation in benchmark datasets including large-scale (up to $9\,\text{km}$) scenarios, where we show state-of-the-art results in VI and VI-LiDAR SLAM.
    \item We will release our framework as open source to foster research in the field of multi-sensor SLAM. 
\end{itemize}

%% file: chapters/02_related-work.tex
\section{Related Work}
\label{sec:related-work}

We review the most relevant works under the topics of multi-sensor SLAM, dense SLAM, and submap-based SLAM.

\subsection{Visual-Inertial SLAM}

VI SLAM systems are generally divided into two main groups: loosely-coupled and tightly-coupled systems. Loosely-coupled estimators typically compute (relative) poses from images alone first before combining with IMU measurements, while the tightly-coupled alternatives consider all measurements simultaneously, typically through IMU kinematics integration with visual observations, e.g.\ in the form of reprojection errors. The latter approach is known to be more accurate. A second distinction of estimator comes from either recursively estimating a selection of recent states employing filtering, e.g.~\cite{msckf} versus optimizing typically a window of states in a nonlinear least-squares fashion, e.g.~\cite{leutenegger2015keyframe, campos2021orb}. On the other hand, depending on the type of residual, direct methods \cite{engel2017direct} minimize the photometric error, while the keypoint-based methods such as~\cite{leutenegger2015keyframe, campos2021orb, VINS-Fusion}, minimize the reprojection error to estimate robot poses and sparse landmarks.

VINS-Mono~\cite{qin2018vins} proposed tightly-coupled nonlinear optimization-based VI SLAM by fusing IMU preintegration and reprojection errors. OpenVINS~\cite{geneva2020openvins} is based on the Multi-State Constraint Kalman Filter (MSCKF~\cite{msckf}), and extends this with long-term SLAM landmarks, extrinsic and intrinsic online calibration. ORB-SLAM3~\cite{campos2021orb} is a tightly-coupled VI-SLAM system that supports multi-session SLAM with monocular, stereo and RGB-D configurations. It records the state-of-the-art accuracy in public SLAM datasets. OKVIS2~\cite{OKVIS2} is a multi-camera SLAM system with loop closure, where posegraph factors are formulated from marginalized landmarks. MAVIS~\cite{wang2024mavis} is also a multi-camera SLAM system with IMU preintegration on matrix Lie Groups. SVIn2 \cite{rahman2019svin2, rahman2022svin2} is a visual-inertial framework that incorporates loop-closures to the factor-graph optimization, while including other sensor modalities for its deployment in underwater environments.

In the context of deep learning-based VI SLAM, DeepVIO~\cite{han2019deepvio} is a self-supervised visual-inertial odometry (VIO) network that directly regresses trajectory from the input sensor measurements. DVI-SLAM~\cite{peng2024dvi} learns a confidence map of feature-metric and reprojection factors with a dense bundle adjustment (DBA) layer. DBA-Fusion~\cite{zhou2024dba} also utilizes the DBA to tightly fuse visual-inertial measurements. The authors demonstrated dense point cloud mapping result in a large-scale environment. However, the DBA layer, which is a main component in end-to-end SLAM systems, often requires high GPU memory usages ($\approx\,24\,\text{GB}$).

In contrast to previous works, OKVIS2-X provides a general framework to work with multiple cameras and multi-modalities including depth networks, LiDAR, and GNSS receiver. Furthermore, our method goes beyond the classical sparse landmark map representation since it can build a dense volumetric occupancy  representation that scales to large scenes.

\subsection{LiDAR(-Visual-Inertial) SLAM}
Due to their inherently accurate depth sensing capabilities, LiDAR sensors have proven to be beneficial for robot localization allowing to reduce the drift significantly, which SLAM systems inevitably suffer from. 
One of the first systems to successfully demonstrate the applicability of LiDAR sensors to SLAM was LOAM~\cite{LOAM}. Its core idea was the extraction of salient geometric features, such as edges and planes, which can be reliably tracked across frames. LOAM has inspired an entire line of research adopting the concept of salient feature extraction~\cite{legoloam,loam_livox}. As in VI SLAM, one can further distinguish feature-based and direct methods. Direct LiDAR SLAM methods, such as Kiss-ICP~\cite{vizzo2023kiss} or CT-ICP~\cite{ct-icp}, typically rely on variants of the Iterative Closest Point (ICP) algorithm on raw LiDAR point clouds, omitting the necessity to extract salient features.
\cite{balm, wiesmann2024efficient} establish sliding-window bundle adjustment (BA) for LiDAR scans for feature-based and direct formulations.

Fusing complementary IMU measurements enables the deployment on highly dynamic systems.
LIO-SAM~\cite{liosam} formulates a tightly-coupled pose graph optimization problem fusing edge and plane error residuals with IMU error residuals. Also filter-based approaches, such as FAST-LIO~\cite{fastlio} or FAST-LIO2~\cite{fastlio2} have successfully demonstrated high-accuracy localization fusing LiDAR and IMU. In contrast to FAST-LIO, which still uses plane and edge features, FAST-LIO2 is a direct method achieving a significant speed-up by directly operating on the raw points.

Recently, LiDAR-Visual-Inertial SLAM systems have become increasingly popular. A large amount of these approaches still make use of LOAM's idea of geometric feature extraction. LVI-SAM~\cite{lvisam} combines a Visual-Inertial (VI) and a LiDAR-Inertial (LI) subsystem to complement each other in challenging scenarios. The LI system builds a factor graph based on IMU preintegration errors and edge and plane residuals.
VILENS~\cite{vilens} also builds an optimisation problem consisting of visual, inertial, robot leg odometry, as well as LiDAR-based line and plane residuals.   
Another line of research uses MSCKF~\cite{msckf}-based approaches for tightly-coupled fusion of visual, inertial and LiDAR measurements, e.g.\ LIC-Fusion~\cite{licfusion} and its successor LIC-Fusion 2.0~\cite{licfusion2}. R2LIVE~\cite{r2live} fuses visual features, IMU and LiDAR measurements in an Iterated Kalman Filter. Unlike previously mentioned LVI SLAM systems, the succeeding R3LIVE~\cite{r3live} applies direct methods, increasing its robustness in texture-less or structure-less environments.
FAST-LIVO~\cite{zheng2022fast} and FAST-LIVO2~\cite{zheng2024fastlivo2} also formulate an Iterated Extended Kalman Filter (IEKF) framework. The IEKF update step consists of two sequential state updates: a point-to-plane based LiDAR update and a sparse direct visual update operating on image patches instead of dense pixels. Both updates are computed in a frame-to-map fashion using a local voxel map. 

In contrast to previous approaches, the proposed OKVIS2-X can fuse LiDAR measurements omitting the necessity to extract salient geometric features or any other correspondences, usually a computationally expensive step. Instead, based on our previous work~\cite{boche2024tightlycoupled}, residuals can be directly derived from dense occupancy maps which are incrementally constructed and are directly usable for downstream tasks. These residuals are added to a factor graph optimization problem.

\subsection{Multi-Sensor SLAM with GNSS fusion}

Early work on fusing global position measurements to reduce the drift in VI SLAM is dominated by filter-based methods, mostly building upon the seminal work in~\cite{msckf} where the authors propose an Extended Kalman Filter (EKF) to tackle realtime visual-inertial navigation.
\cite{lynen2013robust} and~\cite{shen2014multi} use an EKF and an Unscented Kalman Filter (UKF), respectively, to fuse different sensor modalities, such as LiDAR and GNSS measurements. 
Additionally, \cite{lee2020intermittent} also estimates the IMU-GNSS spatial extrinsics as well as the sensor time offset online in an MSCKF~\cite{msckf} framework. \cite{gao2024robust} proposes a robust Iterated Error State Kalman Filter (IESKF) to fuse GNSS, IMU and LiDAR measurements in a tightly-coupled approach. A filter-based system that not only fuses Lidar-Visual-Inertial information, but also wheel odometry and GNSS is MINS \cite{lee2023mins}, a system that extends \cite{geneva2020openvins} and demonstrates how multi-sensor information improves the robustness of state-estimation. Filter-based methods are in general computationally efficient. Nevertheless, as investigated in~\cite{strasdat2010real}, optimization-based approaches potentially deliver superior results, thanks to their ability of re-linearizing old state estimates.

VINS-Fusion~\cite{VINS-Fusion} and GOMSF~\cite{GOMSF} are examples for optimization-based methods fusing global position measurements in a loosely-coupled back-end pose-graph optimization. 
However, a tightly coupled fusion in a unified optimization problem is desirable to fully exploit the available information given by different sensor modalities.
\cite{liosam} offers the integration of GNSS factors into LiDAR-Inertial Odometry as a factor graph optimization problem.
\cite{cioffi2020tightly} proposes a tightly-coupled approach fusing GNSS measurements in the optimization window of a VIO as global factors leveraging IMU preintegration. \cite{liu2021optimization,GVINS,zhang2024gnss} have also proposed tightly-coupled optimization-based frameworks which consider not only global position measurements but also pseudo range and Doppler shift errors from raw GNSS measurements. Unlike the previous, pure odometry approaches,~\cite{cremona2024gnss,li2023lc} successfully demonstrated the fusion of global position measurements into VI SLAM including visual loop closures.

While~\cite{cioffi2020tightly} assumes global position measurements given in the visual-inertial reference frame, in practice, a 4-Degree-of-Freedom (DoF) transformation between a global and the local reference fame has to be estimated. Initialization of this global frame alignment has been addressed using SVD based on correspondences of local and global position measurements~\mbox{\cite{GOMSF}} or in a tightly-coupled fashion using least-squares minimization~\mbox{\cite{lee2020intermittent,liu2021optimization,GVINS}}. \cite{lee2020intermittent} furthermore introduces an heuristic criterion on the observability of the 4-DoF transformation based on the distance traveled. As in~\cite{GOMSF} and following our previous work~\cite{boche2022gnss}, we will use a SVD-based initialization of the GNSS-VIO extrinsics. However, it will be estimated in a tightly-coupled way. In contrast to the aforementioned approaches, to simplify the problem, the global reference frame will be fixed in our approach as soon as it becomes observable. Instead of applying a heuristic threshold as in~\cite{lee2020intermittent}, we introduce a fully uncertainty-aware criterion to determine the covariance of the estimated GNSS extrinsics. Furthermore, our framework supports not only visual loop-closures but also provides a loop-closure like optimization approach to effectively tackle the issue of drift during longer periods of GNSS outage as presented in~\cite{boche2022gnss}. In new experiments, here, we furthermore present how OKVIS2-X supports Lidar-Visual-Inertial SLAM with intermittently available GNSS.

\subsection{Dense SLAM \& Mapping}
Most of the previously mentioned SLAM systems achieve high accuracy in localization relying only on sparse map representations, such as sparse 3D landmarks or feature maps. 
However, understanding dense geometry is indispensable for deploying fully autonomous robots in downstream tasks.

KinectFusion~\cite{newcombe2011kinectfusion} pioneered the area of dense SLAM with Truncated Signed Distance Fields (TSDF) fusion and frame-to-model registration. ElasticFusion~\cite{whelan2016elasticfusion} models an environment with dense surfels, which are updated non-rigidly after a loop. CNN-SLAM~\cite{tateno2017cnn} and DeepFusion~\cite{laidlow2019deepfusion} employed dense depth from a convolutional neural network and fused depths from motion stereo to further improve the depth quality. Kimera~\cite{rosinol2021kimera} and its extension Kimera2~\cite{abate2023kimera2} presented the dynamic scene graph built by visual-inertial observations. In their map representation, semantically annotated meshes are deformed by loop-closure constraints. TANDEM~\cite{koestler2022tandem} integrated a deep Multi-View Stereo (MVS) network in visual odometry, additionally rendering depth from TSDF for image alignment. Simplemapping~\cite{xin2023simplemapping} adopted SimpleRecon~\cite{sayed2022simplerecon} and fused depths from the MVS network into a TSDF given poses and sparse landmarks from ORB-SLAM3~\cite{campos2021orb} but without any feedback from mapping. SigmaFusion~\cite{rosinol2023probabilistic} predicted dense depth uncertainty based on the information matrix from DroidSLAM~\cite{teed2021droid} and showed less noisy reconstruction through uncertainty integration. However, their method is memory intensive, and mapping and state estimation are decoupled. A new line of work based on geometric foundation models has recently emerged in the SLAM community, with MASt3R-SLAM \cite{murai2025mast3r}, a monocular SLAM system, being one of the most notable examples. These systems leverage point maps for both mapping and localization.

In the context of dense neural SLAM, where map geometry and appearance benefits from the neural implicit representation, NICE-SLAM~\cite{zhu2022nice} represents the map with hierarchical feature grids that are decoded into a volumetric occupancy map. UncLe-SLAM~\cite{sandstrom2023uncle} incorporates pixel-wise depth and color uncertainty in tracking and mapping loss functions in NICE-SLAM to properly account for observation uncertainty. Point-SLAM~\cite{sandstrom2023point} refrains from a grid-based representation to save memory by assigning neural points only around surfaces that are later used to obtain the map occupancy and color. Loopy-SLAM~\cite{liso2024loopy} extends Point-SLAM by incorporating loop-closure constraints in the neural point representation by adopting a submap strategy where neural points are anchored at each submap to build a globally consistent map. Aforementioned dense neural SLAM methods have shown promising appearance rendering in (multi-)room-sized environments. However, it is not clear how those lines of works are scalable to km-level large-scale scenarios, which is the main challenge this paper addresses, given the high dimensionality of neural features -- or how these representations efficiently support mobile robot navigation.

In contrast, our proposed system supports tight-coupling between the state estimator and the volumetric free-space mapping for generic sensor setups, as presented in our previous works for depth cameras~\cite{jung2024uncertainty} and LiDAR sensors~\cite{boche2024tightlycoupled}. The proposed approach is fully probabilistic, enabling the consideration of uncertainties also in downstream tasks.

\subsection{Submapping Approaches}
To minimize drift in long-term scenarios, the latest research commonly applies the concept of submapping. This originates from early SLAM research, such as the Atlas framework~\cite{atlas}.
In this context, additional factors can be derived to align submaps and to reduce drift. \cite{vilens} uses local point cloud submaps and adds ICP odometry measurements into the factor graph. Wildcat~\cite{Wildcat}, a sliding-window optimization-based LiDAR-Inertial Odometry system, achieves peak state-of-the-art robustness and accuracy by building local surfel submaps and aligning the submaps. These methods use point or surface-based 3D representations. While these can evidently be leveraged for high-accuracy localization and 3D reconstruction, the map representation is not suitable for navigation due to the difficulty to distinguish between free and unobserved space.

Submapping on volumetric maps has been addressed in various works. \cite{ho2018virtual} builds occupancy submaps based on OctoMap~\cite{octomap} and aligns them by standard ICP registration. Voxgraph~\cite{voxgraph} uses VIO to provide poses for integration into TSDF maps. Upon completion, Euclidean Signed Distance Fields (ESDF) are generated and submaps are aligned in pose graph optimization. New submaps are created at a fixed frequency resulting in a comparably large memory usage. \cite{oxfordLidarSE} instead use occupancy maps as their 3D representation. Using Supereight2~\cite{SE2}, an adaptive-resolution mapping approach, new submaps are spawned based on the distance traveled. Submaps are re-arranged based on updates from the visual-inertial estimator. The follow-up work~\cite{oxfordSubmappingExtended} improved the submap creation by evaluating the point cloud overlaps of new scans and alignment of submaps is based on ICP.

In this work, we will also adopt the concept of submapping. In contrast to~\cite{voxgraph,oxfordLidarSE, oxfordSubmappingExtended}, the global alignment of submaps is not decoupled from the estimator but provides direct feedback. We formulate correspondence-free residuals as in~\cite{voxgraph} without the expensive need to extract ESDFs as we directly use the available occupancy information.

%% file: chapters/03_preliminaries.tex
\section{Preliminaries}
\label{sec:preliminaries}

\subsection{Notation and Definitions}
The classic VI SLAM problem formulation includes several different coordinate frames. A moving body is tracked with respect to a fixed world reference frame $\cframe{W}$. We will denote the IMU frame by $\cframe{S}$ and the camera coordinate frames by $\cframe{C_{i}}$ for $ i = 1\dots N$ cameras. Fusing LiDAR requires introducing the LiDAR sensor frame $\cframe{L}$ and fusing GNSS measurements requires to introduce a global reference frame $\cframe{G}$ which is a gravity-aligned East-North-UP (ENU) local Cartesian frame located in the global position of the first received GNSS measurement. The submap frame is denoted as $\cframe{M}$ that corresponds to $\cframe{S}$ when the new submap is declared.

The rigid body transformation $\T{A}{B} \in SE(3)$ transforms homogeneous points between two frames: $\pos{A}{P} = \T{A}{B}\pos{B}{P}$, where $\pos{A}{P}$ is the position of a point $P$ in frame $\cframe{A}$. The rotational part of $\T{A}{B}$ is expressed by $\mathbf{C}_{AB} \in SO(3)$ and $\pos{A}{B}$ denotes the translation component. We also denote the rotation $\mathbf{C}_{AB}$ with its unit quaternion form $\q{A}{B}$.

\subsection{State Definition}
The state representation in this work is:
\begin{equation}
\label{eq:state-vector}
    \mathbf{x} = \left[ \pos{W}{S}^{T} , \q{W}{S}^{T}, \vel{W}[][]^{T}, \mathbf{b}_\mathrm{g}^{T} , \mathbf{b}_\mathrm{a}^{T} \right]^{T} ,
\end{equation}
where $\pos{W}{S}$, $\q{W}{S}$ and $\vel{W}[][]$ denote the position, orientation and velocity of the IMU sensor frame in the fixed world frame. $\mathbf{b}_\mathrm{g}$ and $\mathbf{b}_\mathrm{a}$ stand for gyroscope and accelerometer biases, respectively. To fuse the visual-inertial system and global measurements, we estimate the extrinsic transformation $[\pos{G}{W}^{T} , \q{G}{W}^{T}]$ between the world reference frame of the estimator $\cframe{W}$ and the GNSS reference frame $\cframe{G}$. In case of online calibration of the camera-IMU extrinsics, the state is further augmented with the extrinsics $[\pos{S}{C_i}^{T} , \q{S}{C_i}^{T}]$ for $i = 1\dots N $ with $N$ being the number of cameras.

%% file: chapters/04_OccMapping.tex
\section{Volumetric occupancy mapping}
\label{sec:mapping}

Obtaining an accurate map representation is essential for state-estimation and downstream tasks. Most of the state-of-the-art VI SLAM systems \cite{geneva2020openvins, campos2021orb, usenko2019visual, wang2024mavis} build a sparse map based on image features, which is used for state estimation but cannot be leveraged for most downstream tasks due to the lack of dense geometric details. To address this, OKVIS2-X leverages dense volumetric occupancy submaps based on the multi-resolution mapping framework Supereight2~\cite{SE2}, which are also considered in state estimation and can be used in downstream tasks such as autonomous navigation \cite{barbas2024digiforest}, exploration \cite{papatheodorou2024efficient} or place recognition \cite{knights2024solvr}.

In OKVIS2-X, we refrain from the concept of monolithic mapping and we leverage a submapping strategy for environment representation, obtaining two major benefits. First, data is allocated into smaller maps, resulting in a shallower data-structure and consequently reducing the computational cost of integrating the depth information. Second, each submap is anchored to individual keyframe states from the state estimator, enabling submap poses to be adjusted during updates. 

Upon creation, the submap reference frame $\cframe{M}$ is anchored to a keyframe state $\mathbf{x}^k$, and its body pose $\T{W}{S_k}$ expresses the relationship between $\cframe{M}$ and $\cframe{W}$. Upon an update of $\mathbf{x}^k$, the new body pose $\T{W}{S_k}$ updates the rigid transformation between $ \cframe{M}$ and $\cframe{W}$, ensuring the local consistency between submaps and improving the overall mapping accuracy. Each submap maintains occupancy log-odds which is updated with either LiDAR point clouds or depth images $\mbf{D}$. The log-odd of a point in the map, $\leftidx{_{M}}{\mathbf{p}}$ is defined as
\begin{equation}
    l\left( \leftidx{_{M}}{\mathbf{p}} \right) = \text{log} \frac{P_{\text{occ}}\left( \leftidx{_{M}}{\mathbf{p}} \:|\: \T{M}{C}, \mbf{D} \right)}{1 - P_{\text{occ}}\left( \leftidx{_{M}}{\mathbf{p}} \:|\: \T{M}{C}, \mbf{D} \right)},
\end{equation}
where $P_\text{occ}$ is the occupancy probability. As in Supereight2, we perform additive Bayesian updates of the log-odds occupancies as we integrate new depth information. The mean of log-odd $L(\cdot)$ is recursively updated as
\begin{align}
    L_k\left( \leftidx{_{M}}{\mathbf{p}} \right) &= \frac{L_{k-1}\left( \leftidx{_{M}}{\mathbf{p}} \right) w_{k-1} + l\left( \leftidx{_{M}}{\mathbf{p}} \right)}{w_{k-1}+1}, \nonumber \\
    w_{k} &= \mathrm{min}  \{ w_{k-1}+1, \: w_\text{max} \}, \label{eq:L}
\end{align}
where $w_k$ is the number of observations and saturated in $w_\text{max}$. This update alleviates the geometric inconsistencies from non-accurate depth perception, more critical when our system leverages learned depth, while also increasing the robustness to the dynamic entities that are perceived from the scene.

The assumption of our submap strategy is that drift within a submap is negligible, therefore to generate a new submap two criteria have to be met. First, a minimum number of depth measurements has been integrated into the current submap. Second, the overlap between the current depth or LiDAR measurement and the previously completed submap is lower than a threshold or that a maximum of $K$ keyframes have been generated since the creation of the current submap. The keyframe criterion is a proxy for the drift within a submap and the overlap criteria ensures that submap-based alignment constraints can reliably be added to the state estimator.

By leveraging occupancy probabilities as a map representation, the environment can be classified as free, occupied or unobserved, a distinction that becomes essential for safe autonomous navigation. Similar to~\cite{barbas2024digiforest}, the path planner only traverses areas considered free in a submap, since it considers unknown areas as non-traversable, and these paths are elastically deformed, according to the submap updates.

Supereight2~\cite{SE2} uses a piecewise linear inverse sensor model for the log-odds occupancy as a function of the measured depth. Hereby, a log-odds value of zero represents the measured surface position. Moreover, the depth uncertainty $\sigma$ can also be modeled as a function of the measured depth $z_r$. In our work, we assume a linearly growing uncertainty for the LiDAR sensor and quadratically for RGB-D cameras.

However, these heuristic uncertainty models are not sufficient for learned depth. For instance, textureless regions or object boundaries yield high depth uncertainty due to the ambiguous correlation volume or viewpoint changes. In response, we will introduce a depth fusion method to predict the pixel-wise uncertainty in the following section.

\begin{figure}
\centerline{\includegraphics[width=\linewidth]{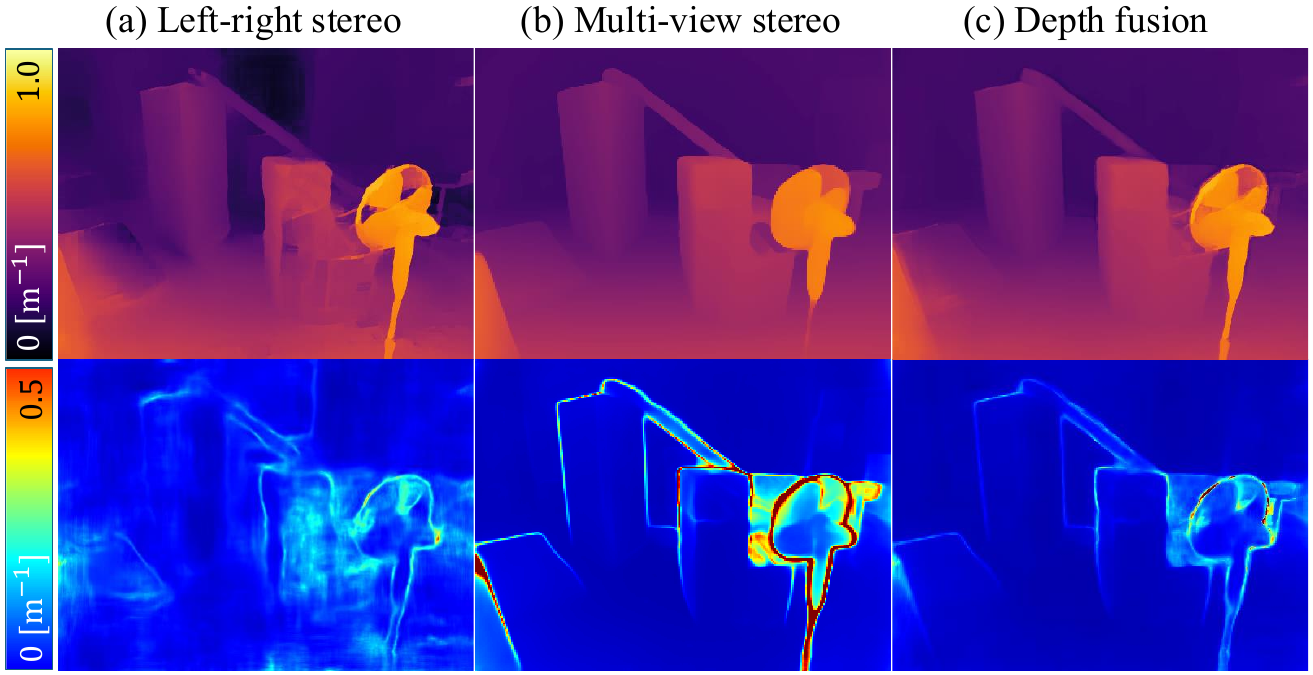}}
\caption{Predicted inverse depth (top) and its corresponding standard deviation (bottom) of (a) stereo network with the $11\,\text{cm}$ baseline, (b) MVS network with the $50\,\text{cm}$ maximum baseline among $8\,$views, and (c) depth fusion in the EuRoC dataset. (Adopted from~\mbox{\cite{jung2024uncertainty}}.)} \label{fig:euroc_images}
\end{figure}

\subsection{Uncertainty-aware depth fusion}
\label{sec:mapping-depth-uncertainty}
It is pivotal to obtain a reliable dense depth as well as the associated uncertainty for our downstream tasks. We follow the key ideas presented in \cite{jung2024uncertainty}, our previous work, where depths from static and motion stereo are fused, which are complementary to each other --- static stereo provides a reliable small baseline even when stationary, while motion stereo potentially brings a large baseline, depending on the camera motion. Our method does not depend on a specific network, but we found that Unimatch~\cite{xu2023unifying} and the MVS network~\cite{xin2023simplemapping} work well in real-world scenes. However, the previous networks only predict disparity or depth without any uncertainties. Therefore, we augment the base architecture with an uncertainty decoder and adopt the Laplacian loss function for (aleatoric) uncertainty learning. Specifically, our stereo network loss function is
\begin{equation}
    \mathcal{L}_{\text{st}}(\mbf{\theta}) = \mathcal{L}_{u}(\mbf{\theta}) + \mathcal{L}_{\nabla u_x}(\mbf{\theta}) + \mathcal{L}_{\nabla u_y}(\mbf{\theta}) \label{eq:stereo_loss},
\end{equation}
where $\mbf{\theta}$ is the network weights, and $u$ stands for the disparity. The disparity loss $\mathcal{L}_u$, modeled in the Laplacian distribution as in~\cite{kendall2017uncertainties, poggi2020uncertainty}, is defined as
\begin{equation}
    \mathcal{L}_{u}(\mbf{\theta}) = \sum_{i\in\mathcal{T}} \frac{\abs{u_i - u_{\text{gt}_i}}}{\sigma_{u_i}} + \log{\sigma_{u_i}},
\end{equation}
where $\mathcal{T}$ is a training set including pairs of stereo images and the ground-truth disparity. We additionally add the gradient loss $\mathcal{L}_{\nabla u}$ for sharper uncertainty output, which is analogously defined as $\mathcal{L}_u$. The gradient uncertainty along the horizontal and vertical directions is derived from the disparity uncertainty as
\begin{align}
    \sigma_{\nabla u_x} &= \sqrt{\sigma^2_{u}(x+1,y) + \sigma^2_{u}(x-1,y)}, \nonumber \\
    \sigma_{\nabla u_y} &= \sqrt{\sigma^2_{u}(x,y+1) + \sigma^2_{u}(x,y-1)}.
\end{align}
We propagate the uncertainty from the disparity to the depth with linearization,
\begin{align}
    \hat{d}_{\text{st}} = \frac{f_c b}{u}, \quad \sigma_{\text{st}} = \frac{f_c b}{u^2} \sigma_u,
\end{align}
where $f_c$ is the rectified focal length, $b$ is the stereo baseline.

Likewise, we modify the loss function of the MVS network~\cite{xin2023simplemapping}
\begin{equation}
    \mathcal{L}_{\text{mvs}}(\mbf{\phi}) = \sum_{i\in\mathcal{T}} \frac{\abs{\log{d_i}-\log{d_{\text{gt}_i}}}}{\sigma_{l_i}} + \log{\sigma_{l_i}},
\end{equation}
where the network learns log-depth uncertainty. We transform the log-depth to a depth with a linearized model,
\begin{align}
    \hat{d}_{\text{mvs}} = \exp{\left(\log{d_i}\right)}, \quad \sigma_{\text{mvs}} = \hat{d}_{\text{mvs}} \sigma_l. \label{eq:mvs_loss}
\end{align}

Given the pixel-wise estimates from the networks $(\hat{d}_\text{st}, \sigma_\text{st})$, $(\hat{d}_\text{mvs}, \sigma_\text{mvs})$ and with the assumption that two estimates are independent, we can optimally fuse two depth estimates~\cite{carlson1990federated},
\begin{align}
    \hat{d}_{\text{fuse}} &= \sigma^2_\text{fuse} \left( \sigma^{-2}_\text{st} \hat{d}_{\text{st}} + \sigma^{-2}_\text{mvs} \hat{d}_{\text{mvs}}\right), \nonumber \\
    \sigma^2_{\text{fuse}} &= \left( \sigma^{-2}_{\text{st}} + \sigma^{-2}_{\text{mvs}}\right)^{-1}. \label{eq:depth_fusion},
\end{align}
Our stereo and MVS networks with additional uncertainty decoders are only fine-tuned in a synthetic dataset~\cite{wang2020tartanair}. Fig.~\ref{fig:euroc_images} shows an example where the fan stand has more detailed depth in the stereo network, while farther objects are sharper in the MVS network. Fused depth naturally maintains the optimal depth based on uncertainty-aware fusion.

%% file: chapters/05_SLAM.tex
\section{Multi-sensor SLAM}
\label{sec:SLAM}

\subsection{System Overview}
\label{sec:SLAM-system-overview}
OKVIS2-X proposes a modular multi-sensor SLAM system, specifically designed for large-scale scenarios and robot navigation. It builds on top of the sparse VI SLAM system OKVIS2~\cite{OKVIS2} and adds capabilities for online camera-IMU extrinsics calibration, fusion of global position measurements and dense map alignment factors derived from LiDAR or depth sensing modalities. Fig.~\ref{fig:system-overview} visualizes the overall system architecture and how it extends the original OKVIS2.

The underlying VI SLAM system is split into the visual frontend and a realtime estimator that process images and IMU messages synchronously whenever a new (multi-) frame arrives. To deal with loop-closures, a full factor graph loop optimization is executed asynchronously. As shown in Fig.~\ref{fig:system-overview}, the frontend deals with state initialization, keypoint matching, stereo triangulation (of successive frames and from stereo images of the same multi-frame), if enabled, running a segmentation CNN (to filter out observations in the sky), as well as place recognition and, if the latter was successful, relocalization and loop-closure initialization. 
The realtime estimator will then optimize the respective factor graph, and is also responsible for the creation of posegraph edges by marginalizing old observations, as well as for fixation of old states. Upon loop-closure, it turns posegraph edges back into observations, and then triggers the optimization of the full graph around a loop, which runs asynchronously, and which will be synchronized with the realtime factor graph upon completion.

OKVIS2-X extends this system by three additional modules: \textit{Depth Network}, \textit{Multi-Sensor Processor}, and the \textit{Submapping Interface}. \textit{Depth Network} takes stereo images for the stereo network, as well as robot poses and sparse landmarks from the realtime estimator for the MVS network. The sparse landmarks are back-projected to image planes to provide a depth prior for the network. Depending on the use case, the \textit{Multi-Sensor Processor} deals  with incoming GNSS measurements and geometric measurements coming either from a LiDAR, a depth camera or the presented depth fusion network. For GNSS measurements, it will formulate global position residuals that are added to the realtime estimator. It further determines the time that has passed since the last received GNSS measurement. In case of long signal outages, it will trigger an asynchronous, loop-closure like optimization of the full graph to compensate for the potentially accumulated drift since the last received measurement. In case of dense submap alignment, it retrieves poses for depth images or applies IMU-based motion undistortion of incoming LiDAR point clouds. For that it keeps an internal representation of the trajectory which is always kept synchronized with the estimator. For every live frame, frame-to-map factors with respect to a previous submap are added to the factor graph.

The \textit{Submapping Interface} manages the collection of all submaps. The pose of each submap is anchored to a keyframe pose from the state estimator, therefore submap poses are always updated with updates from the state estimator using the same internal trajectory representation as the \textit{Multi-Sensor Processor}. Each time a visual keyframe is generated, it checks for the overlap ratio of incoming measurements with respect to the last submap to decide whether a new submap is created. After that decision, depth images or LiDAR measurements are integrated into the current active submap. Upon completion of a submap, dense map-to-map factors are added in the optimization problem to keep consistency of submaps in overlapping regions. In the following sections, we will describe each module in more detail.

\subsection{Visual Frontend}
The visual frontend extracts BRISK~\cite{brisk} keypoints and descriptors in every image of a multi-frame, and matches them to the 3D landmarks already in the map; hereby both, descriptor distance and reprojected image distance, are considered. New 3D landmarks are then initialized both from stereo triangulation within all images of a keyframe, as well as from triangulation between live frame and any of the keyframes. The decision of whether a new frame is considered a keyframe is taken depending on the fraction of matched landmarks in the live frame, as well as the fraction of current matches visible in any of the existing keyframes. If the overlap falls below a threshold, we set the live (multi-)frame as a new keyframe.

A finetuned version of the semantic segmentation network Fast-SCNN~\cite{poudel2019fast} can be run on keyframe images. For maximum portability and flexibility, inference is carried out asynchronously on the CPU, which is tractable, due to the efficient network as well as the fact that only keyframes are processed. In our implementation, matches that are in the sky are removed.
This scheme significantly improves accuracy in presence of slow-moving scene content, particularly clouds, where observations are not already automatically discarded via the Cauchy robustification. 

\subsection{Place Recognition, Relocalization, and loop-closure}
OKVIS2-X maintains a DBoW2~\cite{dbow2} database. DBoW queries of the current frames return a list of matches as loop-closure candidates. To be considered a valid loop-closure, an additional geometric verification step using 3D-2D RANSAC has to be passed. Then, the posegraph edges connecting the loop-closure state will be “revived” and turned back into landmarks and observations (see Fig.~\ref{fig:factor-graph_vi} (c)); and observations with the current matching frame will also be created. This may also trigger merging of landmarks, if already existing new landmarks are matched to old landmarks. To optimize the loop inconsistency, the error is equally distributed around the loop using rotation averaging followed by position inconsistency distribution. Then, a background optimization process of the full graph is triggered. After the optimization has finished, a synchronization process imports the optimized states and landmarks into the realtime estimator, and re-aligns new states and landmarks created in the meantime.

\subsection{Visual-Inertial SLAM}
\begin{figure}[t]
    \centering
    \includegraphics[width=0.95\linewidth]{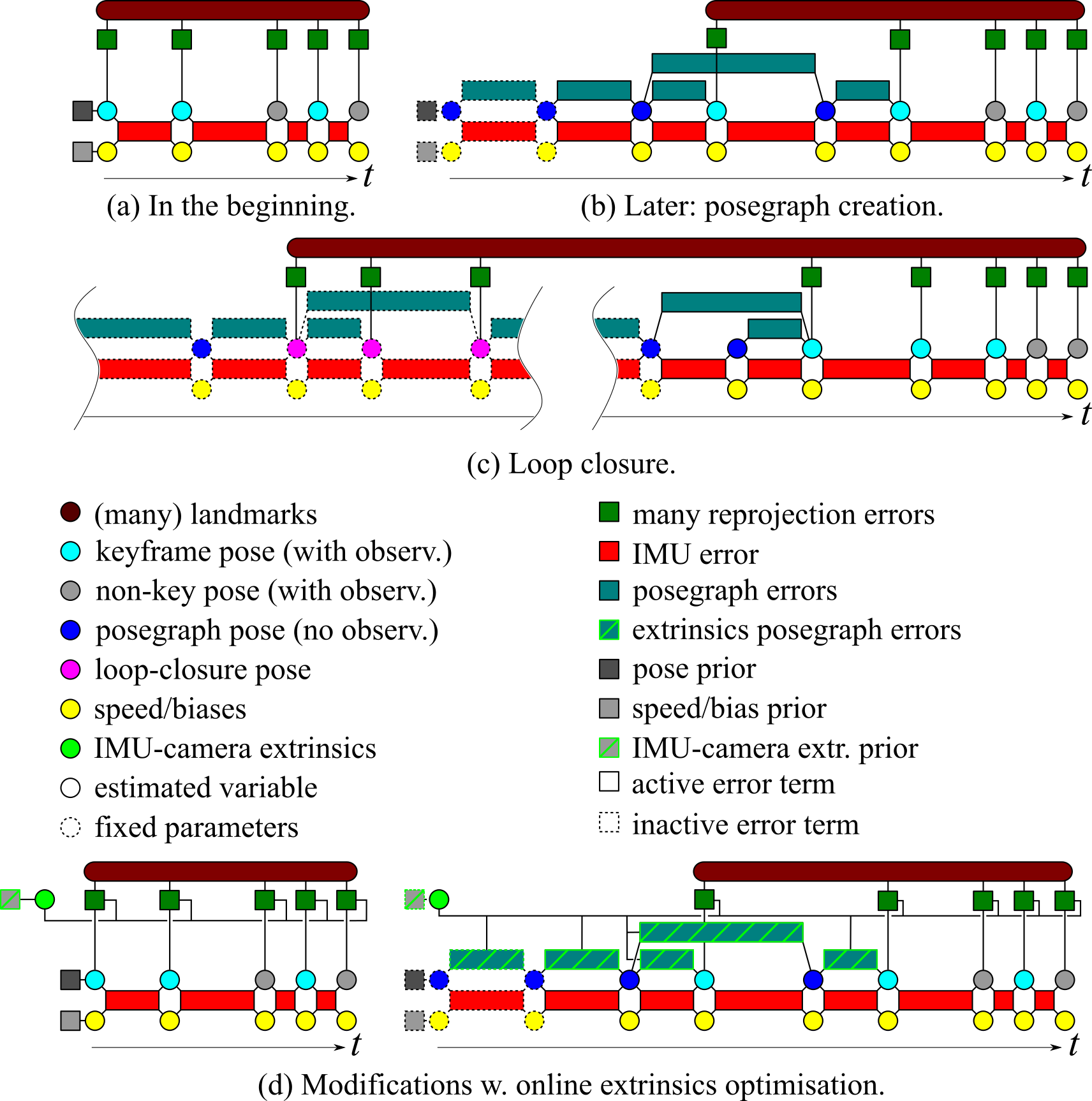}
    \caption{Initially a full batch VI factor graph (a) is created and optimized. Later (b), frames with least overlap with the live frame and current keyframe are turned into posegraph poses by construction of relative pose errors under marginalization of common observations; also, old poses and speed/bias variables are fixed to keep the problem realtime capable. When a loop-closure occurs (c), respective observations and landmarks are re-activated. The proposed system furthermore supports online calibration of the IMU-camera extrinsics (d). ((a-c) Adopted from \cite{OKVIS2}.)}
    \label{fig:factor-graph_vi}
\end{figure}

The VI estimator will be minimizing visual, inertial, and relative pose errors (posegraph edges), which are briefly introduced in what follows. Fig.~\ref{fig:factor-graph_vi}(a)-(c) visually overviews the construction of the underlying factor graph as time progresses.

\subsubsection{Visual Reprojection Errors}

Just like OKVIS2~\cite{OKVIS2}, OKVIS2-X uses reprojection errors $ \mathbf{e}_{\mathrm{r}}^{i,j,k} $ of the $j$-th landmark in the frame of the $i$-th camera at a timestamp $k$ and its corresponding observation $\Tilde{\mathbf{z}}_{\mathrm{r}}^{i,j,k}$ in the image is given by
\begin{equation}
    \label{eq:residuals_reprojection}
    \mathbf{e}_{\mathrm{r}}^{i,j,k} 
    =
    \Tilde{\mathbf{z}}_{\mathrm{r}}^{i,j,k}  - \mathbf{h} \left( \mbfh{T}_{SC_i}^{-1} \mbfh{T}_{WS_k}^{-1} \leftidx{_W}{\mathbf{l}^{j}} \right) .
\end{equation}
Hereby, $\mathbf{h} \left( \cdot \right)$ denotes the camera projection.

\subsubsection{IMU Errors}

For the formulation of the IMU residuals, we adopt the IMU preintegration approach in~\cite{forster2016manifold}. Between time steps $k$ and $n$, the IMU error is:
\begin{equation}
    \label{eq:residuals_imu}
    \mathbf{e}_{\mathrm{s}}^{k} 
    =
    \Tilde{\mathbf{x}}^{n} \left( \mathbf{x}^{k} , \Tilde{\mathbf{z}}_{\mathrm{s}}^{k,n}\right) \boxminus \mathbf{x}^{n} ,
\end{equation}
where $\Tilde{\mathbf{x}}^{n}$ is the predicted state at an arbitrary time $n$ as a function of the current state $ \mathbf{x}^{k} $ and IMU measurements $\Tilde{\mathbf{z}}_{\mathrm{s}}^{k,n}$. The $\boxminus$ performs regular subtraction except for the quaternion (see~\cite{OKVIS2}).

\subsubsection{Relative Posegraph Errors}
Relative pose errors $\mathbf{e}_{\mathrm{p}}^{r,c}$ between time steps $r$ (the reference) and $c$ are given by 
\begin{equation}
    \label{eq:residuals_rel_pose}
    \mathbf{e}_{\mathrm{p}}^{r,c} 
    =
    \mathbf{e}_{\mathrm{p},0}^{r,c}
    +
    \begin{bmatrix}
      \pos{S_r}{S_c} - \posbar{S_r}{S_c} 
      \\
      \mathbf{q}_{S_r S_c} \boxminus {\mbfbar{q}}_{S_r S_c},
    \end{bmatrix}
\end{equation}
with $\posbar{S_r}{S_c}$ and ${\mbfbar{q}}_{S_r S_c}$ being nominal relative position and orientations expressed in the IMU frame $\cframe{S_r}$. 

In short, these posegraph error terms are computed from joint observations into frames $r$ and $c$ by marginalizing out the landmarks.
We refer the reader to~\cite{OKVIS2} for a detailed explanation of how a Maximum Spanning Tree (MST) is employed to select posegraph edges to be created. To compute the constant $\mathbf{e}_{\mathrm{p},0}^{r,c}$ as well as the weight matrix $\mathbf{W}_\mathrm{s}^{k}$, we transform the landmarks into the reference frame $\cframe{S_r}$ and formulate the Gauss-Newton-System of the form $\mbf{H}\delta \mbf{\chi} = \mbf{b}$ just from joint observations and respective landmarks, i.e.\ from standard reprojection errors and respective well-known Jacobians, resulting in the following structure:
\begin{equation}
    \begin{bmatrix}
    \mbf{H}_{\mathrm{p},\mathrm{p}} 
    & \ldots & \mbf{H}_{\mathrm{p},j} & \ldots\\
    \vdots &\ddots &\mbf{0} &\mbf{0}\\
    \mbf{H}_{\mathrm{p},j}^T &\mbf{0} &\mbf{H}_{j,j} &\mbf{0}\\
    \vdots &\mbf{0} &\mbf{0} &\ddots\\
    \end{bmatrix} 
    \begin{bmatrix}
    \delta \mbf{p}\\
    \vdots\\
    \delta \mbf{l}_j \\
    \vdots
    \end{bmatrix}
    = 
    \begin{bmatrix}
    \mbf{b}_{\mathrm{p}} \\
    \vdots\\
    \mbf{b}_{j}\\
    \vdots
    \end{bmatrix}.
\end{equation}
Hereby, only one pose occurs, i.e.\ the relative pose $\T{S_r}{S_c}$ referred to by $\delta \mbf{p}$. The variable ordering follows with all the landmarks (${_{S^r}}\mbf{l}^j$) referred to as $\delta \mbf{l}_j$. Now, landmarks are marginalized out using the Schur complement:
\begin{eqnarray}
\label{e:schur}
\mbf{H}^* &= &\mbf{H}_{\mathrm{p},\mathrm{p}} 
- \sum_j \mbf{H}_{\mathrm{p},j} \mbf{H}_{jj}^+ \mbf{H}_{\mathrm{p}, j}^T,\\
\mbf{b}^* &= &\mbf{b}_\mathrm{p} 
- \sum_j \mbf{H}_{\mathrm{p},j} \mbf{H}_{jj}^+ \mbf{b}_j,
\end{eqnarray}
which yields the reduced Gauss-Newton system
\begin{equation}
    \mbf{H}^* \delta \mbf{p} = \mbf{b}^*,
\end{equation}
which is now linearized around the current pose $\mbfbar{p}$, therefore 
\begin{equation}
\label{e:reduced_GN}
    \mbf{H}^* \delta \mbf{p} = \mbf{b}^* - \mbf{H}^* (\mbf{p}\boxminus\mbfbar{p})
\end{equation}
The supposedly equivalent Gauss-Newton system of a respective posegraph error term takes the form
\begin{equation}
    {\mbf{E}_\mathrm{p}^{r,c}}^T \mbf{W}_\mathrm{p}^{r,c}\mbf{E}_\mathrm{p}^{r,c} \delta \mbf{p} = -{\mbf{E}_\mathrm{p}^{r,c}}^T \mbf{W}_\mathrm{p}^{r,c}(\mbf{e}_\mathrm{p,0}^{r,c}+\mbf{p}\boxminus\mbfbar{p}),
\end{equation}
where $\mbf{E}_\mathrm{p}^{r,c}$ denotes the Jacobian. For this to be equivalent to (\ref{e:reduced_GN}) at the linearization point (i.e.\ at $\mbf{p}=\mbfbar{p}$ where $\mbf{E}_\mathrm{p}^{r,c}=\mbf{I}_6$), we now obtain
\begin{eqnarray}
\mbf{W}_\mathrm{p}^{r,c} = \mbf{H}^*,\\
\mbf{e}_{\mathrm{p},0}^{r,c} = -{\mbf{H}^*}^+ \mbf{b}^*.
\end{eqnarray}
Since $\mbf{H}^*$, $\mbf{b}^*$, and $\mbfbar{p}$ are constants, the Jacobians w.r.t.\ the poses at steps $r$ and $c$ are fairly straightforward to determine after substituting $\T{S_r}{S_c} = \mbfh{T}_{WS_r}^{-1}\T{W}{S_c}$.

\subsection{Online Camera-IMU Extrinsics Calibration}
We have extended \cite{OKVIS2} to fully support online calibration of camera-IMU extrinsic poses $\T{S}{C_i}, i=\{1, \ldots, N\}$. While this extension is straightforward and well-known for reprojection errors, the relative posegraph factors from Eqn.~(\ref{eq:residuals_rel_pose}) now also depend on $\T{S}{C_i}$. Consequently, before marginalizing co-visible landmarks, the two-view Gauss-Newton system will be augmented to include extrinsics poses -- which remain after marginalizing out landmarks. Eqns~(\ref{e:schur}) ff.\ remain the same, but with a higher-dimensional $\mbf H$ and $\mbf{b}$. Eqn.~(\ref{eq:residuals_rel_pose}) is then consequently extended as
\begin{equation}
    \label{eq:residuals_rel_pose_onlinecalib}
    \mathbf{e}_{\mathrm{po}}^{r,c} 
    =
    \mathbf{e}_{\mathrm{po},0}^{r,c}
    +
    \begin{bmatrix}
      \pos{S_r}{S_c} - \posbar{S_r}{S_c} \\
      \q{S_r}{S_c} \boxminus \qbar{S_r}{S_c}\\
      \pos{S}{C_1} - \posbar{S}{C_1}\\
      \q{S}{C_1} \boxminus \qbar{S}{C_1}\\
      \vdots\\
      \pos{S}{C_N} - \posbar{S}{C_N}\\
      \q{S}{C_N} \boxminus \qbar{S}{C_N}
    \end{bmatrix} \in \mathbb{R}^{6+6N},
\end{equation}
where $\posbar{S}{C_i}, \qbar{S}{C_i}$ denote the linearization point of the $i^\text{th}$ extrinsics, i.e.\ their values upon construction of the (augmented) relative pose factor.

The related factor graph structure can be observed in Fig.~\ref{fig:factor-graph_vi}(d). 
Furthermore, to regularize the estimation problem, we add a pose prior to all camera extrinsics using the same formulation as with the pose prior of the estimated robot state\footnote{In case we run a full-batch final BA including extrinsics, this prior is removed beforehand.}.

\begin{figure}[t]
    \centering
    \includegraphics[width=0.9\linewidth]{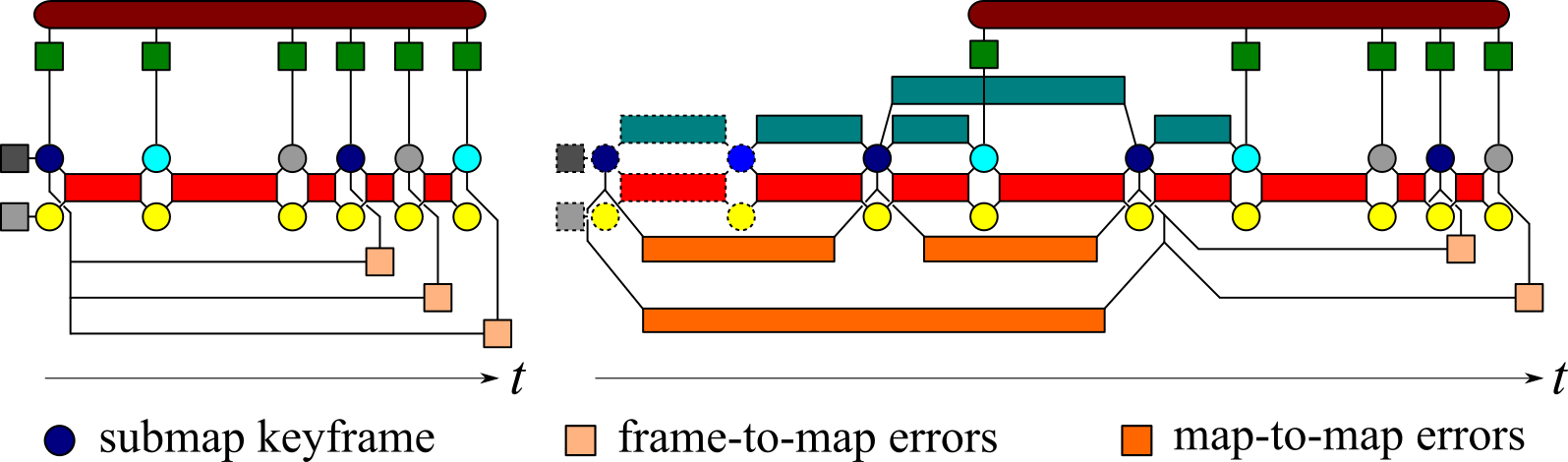}
    \caption{Factor Graph including dense submap alignment. Left: The realtime estimator connects set of current keyframe and non-keyframe states by IMU errors and visual reprojection errors. For every state in the optimization window, frame-to-map factors are formulated between every live state and the keyframe state associated to the last completed submap. Right: Measurements between frames can be aggregated and map-to-map factors can be added to the factor graph between submap keyframe states if the geometric overlap surpasses a threshold. (Adopted from \cite{boche2024tightlycoupled}.)}
    \label{fig:factor-graph_map_align}
\end{figure}

\subsection{Submap Alignment}
As stated in Sec.~\ref{sec:SLAM-system-overview} and visualized in Fig.~\ref{fig:factor-graph_map_align}, there are two types of dense submap alignment residuals. Frame-to-map factors can be formulated between every live frame and the keyframe associated to the last completed submap. Map-to-map residuals are computed between two keyframes that two submaps are anchored to. However, regarding the actual error formulation, they do not differ.
Given a completed submap associated to a keyframe pose $\T{W}{S_a}$ and a point cloud $\mathcal{P}_{b}$ associated to another state $\T{W}{S_b}$, we formulate the map-based residual for every $\mvec{S_b}{p} \in \mathcal{P}_{b}$ as:
\begin{equation}
    \label{eq:method_lidar_residual}
    e^{a,b}_{\mathrm{m}} \left( \mvec{S_a}{p} \right) =
    \frac{d}{\sigma} 
    = 
    \frac{L\left( \mvec{S_a}{p} \right)}
    {\sqrt{\frac{L_{\mathrm{min}}^{2}}{9} + \sigma_{d}^{2} \left| \nabla L \left( \mvec{S_a}{p} \right)\right|^{2}}} ,
\end{equation}
where $\mvec{S_a}{p} = \T{S_a}{S_b} \mvec{S_b}{p}$ with $\T{S_a}{S_b} = \T{W}{S_a}^{-1} \T{W}{S_b}$. The respective Jacobians can be found in Appendix~\ref{appdx:submap-alignment-jacobian}. In the case of depth images, points $\mvec{S_b}{p}$ are obtained by backprojecting pixel depth values. The idea of the map alignment residuals is that every measured point should be on a surface ($L=0$) in the 3D map; and the distance $d$ of the point from the nearest surface can be extrapolated from the occupancy value $L\left( \cdot \right)$ and the occupancy gradient $\nabla L \left( \cdot \right) $ assuming a linear behavior near the surface as in the sensor model for mapping. The distance $d$ and the map uncertainty can be derived as:
\begin{equation}
    \label{eq:method_map_distance_uncertainty}
    d = \frac{L}{|\nabla L|}, \qquad
    \sigma_{\mathrm{map}} = \frac{L_{\mathrm{min}}}{3 | \nabla L |} .
\end{equation}
$L_{\mathrm{min}}$ is a configuration parameter and denotes the saturation minimum log-odds occupancy value.
With the sensor-specific measurement uncertainty $\sigma_{d}$, we can formulate the weighted residual in Eqn.~\eqref{eq:method_lidar_residual} using the total uncertainty
\begin{equation}
    \label{eq:method_total_uncertainty}
    \sigma = \sqrt{\sigma_{\mathrm{map}}^{2} + \sigma_{d}^{2}}.
\end{equation}
While we found that for the typically highly precise LiDAR measurements, the simple assumption of an isotropic sensor uncertainty is sufficient, this does not hold for depth from the network. 
In that case, the depth uncertainty as in Eqn~(\ref{eq:depth_fusion}), which often includes noisy observations, should be assigned pixel-wise following the approach descriped in Sec.~\ref{sec:mapping-depth-uncertainty} to properly weight its contribution to the optimization loss.

\subsection{GNSS Fusion}
In order to fuse global position residuals, such as from GNSS, we augment the state vector to also contain the 4-DoF transformation $\T{G}{W}$ between the VI reference frame $\cframe{W}$ and a global reference frame $\cframe{G}$ which is a gravity-aligned East-North-Up (ENU) local Cartesian frame at the position of the first measurement.
\subsubsection{Global Position Errors}

\begin{figure}[t]
    \centering
    \includegraphics[width=0.8\linewidth]{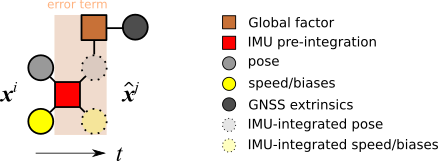}
    \caption{To evaluate the global position residual, the propagated state $\hat{\mathbf{x}}\attime{j}$ at time step $j$ of the measurement is propagated from the state vector $\mathbf{x}\attime{i}$ at time step $i$ of the camera frame by leveraging IMU preintegration. This intermediate state is only used for evaluation of the global factor. (Adopted from \cite{boche2022gnss}.)}
    \label{fig:factor-graph_vig}
\end{figure}

The global position residual for a measurement $\mathbf{z}\attime{j}$ at a time step $j$ can be formulated as:
\begin{equation}
\label{eq:gps res}
    \mathbf{e}_\mathrm{g}^{j} = \mathbf{z}\attime{j} - 
    \left[\C{G}{W} \left( \poshat{W}{S_j} + \hat{\mathbf{C}}_{WS_j} \pos{S}{A} \right) + \pos{G}{W} \right].
\end{equation}
Hereby, $\pos{S}{A}$ considers the position of the GNSS antenna in the IMU sensor frame which is assumed to be known beforehand. To account for asynchronous arrival of GNSS measurements and camera frames, $\poshat{W}{S_j}$ and $\hat{\mathbf{C}}_{WS_j}$ are predictions of the IMU poses at time step $j$ which can be obtained from a previous state at time step $i$ through IMU preintegration. This is visualized in Fig.~\ref{fig:factor-graph_vig}. Compared to~\cite{cioffi2020tightly}, the formulation in~\eqref{eq:gps res} extends the global residual by also considering a 4 DoF transformation $\T{G}{W}$ between the global and a local reference frame. The error Jacobians are derived in Appendix~\ref{appdx:gnss-jacobians}.

\subsubsection{Global Reference Frame initialization}
\label{sec:SLAM-gnss-init}
A reliable initialization of the global reference frame with respect to the VI reference frame is crucial. It is desirable to define an uncertainty-aware criterion for the observability of the 4-DoF extrinsic transformation between the two reference frames based on the received measurements. 
In a first step, an initial solution can be obtained using the SVD-based alignment method presented in~\cite{posyaw} from correspondences of global measurements and poses in the world reference frame. 

Subsequently, the reliability of this initial solution can be evaluated. For $N_\mathrm{g}$ received measurements, global position errors $ \mathbf{e}_\mathrm{g}^{j}$ ($j=1,\dots,N_\mathrm{g}$) and the corresponding error Jacobians $\mathbf{E}_\mathrm{g}^{j}$ are computed based on Eqn~\eqref{eq:gps res}. 
The covariance matrix $\mathbf{P}$ for the 4 DoF transformation can be estimated as the inverse of the approximate Hessian $\mathbf{H}$:
\begin{equation}
\label{eq:init_hessian}
    \mathbf{H} = \sum_{j=1}^{N_\mathrm{g}} \left( {\mathbf{E}_\mathrm{g}^{j}}^{T} \mathbf{W}_\mathrm{g}\attime{j} \mathbf{E}_\mathrm{g}^{j} \right) ,
\end{equation}
where $\mathbf{W}_\mathrm{g}\attime{j}$ is the inverse of the overall covariance matrix $\boldsymbol{\Sigma}_\mathrm{g}\attime{j}$ for GNSS residuals, which considers GNSS measurement covariances as well as IMU preintegration covariances (see Appendix~\ref{appdx:gnss-covariance} for more details). 
We examine the variance $p_{\theta \theta} $ corresponding to the yaw angle $\theta$ in $\mathbf{P}$, and define a decision threshold $\sigma_{\theta}^{2}$ on whether the global reference frame is observable as $p_{\theta \theta} < \sigma_{\theta}^{2}$.
When the yaw uncertainty falls below this threshold, $\T{G}{W}$ can be assumed to be known and fixed.

\subsubsection{GNSS Alignment}
\label{sec:SLAM-gnss-align}
This mechanism of initializing and fixing the GNSS extrinsics enables a global alignment strategy to compensate for drift accumulated throughout potential GNSS signal dropouts.
OKVIS2-X limits the computational complexity of the optimization problem by fixing states that are far in the past. Whenever GNSS residuals are added to the graph optimization problem, we check the fixation status of the last state that has an associated global position factor. Dropouts in receiving GNSS signals over a long period of time can be identified by whether the last state with a GNSS factor in the graph is already frozen. In that case, the trajectory might have accumulated a significant amount of drift since then.
Eliminating this drift can be addressed in a loop-closure like global alignment approach. After detecting a dropout, a re-initialization of the GNSS extrinsics is done as described in the previous paragraph.
From this re-initialization process, a new estimate for the extrinsic transformation, $\T{G}{W_\mathrm{new}}$, is obtained. The discrepancy between the originally estimated $\T{G}{W}$ and the newly initialized $\T{G}{W_\mathrm{new}}$ can be calculated as:
\begin{equation}
    \T{W_\mathrm{new}}{W} = \mbfh{T}^{-1}_{GW_\mathrm{new}}\hspace{1mm}\T{G}{W},
\end{equation}
which also gives an estimate for the trajectory drift. Using this delta transformation and rotation averaging, the position and orientation error can be distributed across all the states during a GNSS dropout. After this alignment, an optimization of the full graph is triggered.

\subsection{Factor Graph optimization}
All of the aforementioned factors are combined in the overall minimization objective:
\begin{align}
    \label{eq:optimization_objective}
    c\left( \mathbf{x} \right) &= 
    \frac{1}{2} \sum_{i} \sum_{k \in \mathcal{K}} \sum_{j \in \mathcal{J} \left(i,k \right)} \rho_{\mathrm{c}} \left( {\mathbf{e}_{\mathrm{r}}^{i,j,k}}^{T} \mathbf{W}_{\mathrm{r}} \mathbf{e}_{\mathrm{r}}^{i,j,k}   \right)
    \nonumber \\
    &+ \frac{1}{2} \sum_{k \in \mathcal{P} \cup \mathcal{K} \setminus f} 
    {\mathbf{e}_{\mathrm{s}}^{k}}^{T} \mathbf{W}_\mathrm{s}^{k} \mathbf{e}_{\mathrm{s}}^{k} 
    + \frac{1}{2} \sum_{r\in \mathcal{P}} \sum_{c\in \mathcal{C}\left(r\right)} {\mathbf{e}_{\mathrm{p}}^{r,c}}^{T} \mathbf{W}_{\mathrm{r}}^{r,c} \mathbf{e}_{\mathrm{p}}^{r,c}   
    \nonumber \\
    &+ \frac{1}{2} \sum_{k \in \mathcal{K}}\sum_{\mathbf{p} \in \mathcal{L}_k}{\rho_{\mathrm{t}} \left({e_{\mathrm{m}}^{C,k}}^2 \right)}
    + \frac{1}{2} \sum_{b \in \mathcal{M}}\sum_{a \in \mathcal{A}_b}\sum_{\mathbf{p} \in \mathcal{L}_b}{ \rho_{\mathrm{t}} \left( {e_{\mathrm{m}}^{a,b}}^2\right)} 
    \nonumber \\
    &+ \frac{1}{2} \sum_{j \in \mathcal{G}}{{\mathbf{e}_\mathrm{g}^{j}}^{T} \mathbf{W}_\mathrm{g}\attime{j} \mathbf{e}_\mathrm{g}^{j}}.
\end{align}
Here, the set $\mathcal{K}$ contains the most recent frames as well as keyframes with observations of visible landmarks in $\mathcal{J} \left(i,k \right)$. $\mathcal{P}$ contains all posegraph frames and $f$ denotes the most current frame. $\mathcal{C}\left(r\right) \subset \mathcal{P}$ is the set of all posegraph frames connected to a frame $r$. 
Furthermore, the set $\mathcal{L}_k$ denotes the set of all map alignment residuals associated to a frame $k$. $\mathcal{M}$ is the set of all past submaps, and $\mathcal{A}_b$ the set of all submap frames connected to a submap frame $\T{W}{S_b}$ via map-to-map residuals. $C$ denotes the last completed submap frame. $\mathcal{G}$ is the set of added GNSS residuals. The Cauchy robustifier $\rho_{\mathrm{c}} \left(\cdot\right)$ and Tukey robustifier $\rho_{\mathrm{t}} \left(\cdot\right)$ are used for reprojection errors and map alignment errors.

%% file: chapters/06_evaluation.tex
\setlength{\tabcolsep}{6.0pt}
\begin{table}[tb]
\begin{center}
\renewcommand{\arraystretch}{1.1} 
\caption{Method definition based on a sensor configurations and estimator causality}
\label{tab:definition_variations}
\begin{tabular}{l | c c c c c | c c c}
\Xhline{3\arrayrulewidth}
 & \multicolumn{5}{c|}{\textbf{Sensors}} & \multicolumn{3}{c}{\textbf{Causality}} \\ \hline
 \textbf{Methods} & \rotatebox[origin=c]{90}{\textit{Visual}} & \rotatebox[origin=c]{90}{\textit{Inertial}} & \rotatebox[origin=c]{90}{\textit{Depth}} & \rotatebox[origin=c]{90}{\textit{LiDAR}} & \rotatebox[origin=c]{90}{\textit{GNSS}} & \rotatebox[origin=c]{90}{\textit{Causal}} & \rotatebox[origin=c]{90}{\textit{Non-causal}} & \rotatebox[origin=c]{90}{\textit{Full-BA}}  \\
\hline

\textit{Ours-vi-c} & \checkmark & \checkmark & & & & \checkmark & & \\
\textit{Ours-vi-nc} & \checkmark & \checkmark & & & & & \checkmark & \\
\textit{Ours-vi-ba} & \checkmark & \checkmark & & & & &(\checkmark) & \checkmark \\
\textit{Ours-v-}$\star$ &\checkmark & & & & & $\star$ & $\star$ & $\star$ \\
\textit{Ours-vig-}$\star$ &\checkmark & \checkmark & & & \checkmark & $\star$ & $\star$ & $\star$ \\
\textit{Ours-vid-}$\star$ & \checkmark & \checkmark & \checkmark & & & $\star$ & $\star$ & $\star$ \\
\textit{Ours-vidg-}$\star$ &\checkmark & \checkmark & \checkmark & & \checkmark & $\star$ & $\star$ & $\star$ \\
\textit{Ours-vil-}$\star$ &\checkmark & \checkmark & & \checkmark & & $\star$ & $\star$ & $\star$ \\
\textit{Ours-vilg-}$\star$ &\checkmark & \checkmark & & \checkmark & \checkmark & $\star$& $\star$& $\star$\\
\Xhline{3\arrayrulewidth}
\end{tabular}
\end{center}
\end{table}

\section{Evaluation}
\label{sec:eval}
The objective of this evaluation is to quantify the trajectory accuracy and mapping accuracy and completeness of OKVIS2-X with respect to state-of-the-art methods in small to large-scale scenarios. The large-scale environment (km-level traveling distance) possesses extreme challenges due to huge drift over time and memory usage in dense mapping to cover the large area. To showcase OKVIS2-X under such environments, we selected three public datasets: EuRoC~\cite{Euroc}, Hilti-Oxford~\cite{Hilti22}, and VBR~\cite{VBR} datasets where the traveled distance spans from tens of meters with a flying drone (EuRoC), to hundreds of meters with a hand-held sensor rig (Hilti-Oxford) to $9 \, \text{km}$ for a driving car (VBR). For the proposed GNSS fusion, we further added an evaluation in a challenging sequence from the GVINS dataset\mbox{\cite{GVINS}} including indoor-outdoor transitions and cluttered environments. All evaluations were performed in a serial manner, ensuring no frame drops. We account for the time offset between camera and IMU, if known (e.g. in the Hilti-Oxford dataset), as well as time offsets for both, LiDAR and GNSS, with respect to the IMU.

\setlength{\tabcolsep}{3.0pt}
\begin{table*}[tb]
\caption{Root mean square of absolute trajectory error [meter] in the EuRoC dataset}
\label{tab:euroc_ate}
\begin{center}
\begin{threeparttable}[b]
\renewcommand{\arraystretch}{1.4} 
\begin{tabular}{c | c | l | c c c c c c c c c c c | c}
\Xhline{3\arrayrulewidth}
\multicolumn{2}{c|}{} & \multirow{2}{*}{\textbf{Methods}} & \multicolumn{11}{c|}{\textbf{Sequence}} & \multirow{2}{*}{\textbf{Avg}} \\ \cline{4-14}
\multicolumn{2}{c|}{} & & \texttt{MH01} & \texttt{MH02} & \texttt{MH03} & \texttt{MH04} & \texttt{MH05} & \texttt{V101} & \texttt{V102} & \texttt{V103} & \texttt{V201} & \texttt{V202} & \texttt{V203} \\
\Xhline{3\arrayrulewidth}

\multirow{4}{*}{\rotatebox[origin=c]{90}{V-SLAM}} & \rotatebox[origin=c]{90}{C} & \textit{Ours-v-c} & 0.034 & 0.038 & 0.048 & 0.122 & 0.123 & 0.040 & 0.054 & 0.123 & 0.072 & 0.095 & 1.133\tnote{4} & 0.075 \\ \cline{2-15}

& \multirow{3}{*}{\rotatebox[origin=c]{90}{NC}} & ORB-SLAM3~\cite{campos2021orb}\tnote{1} & 0.029 & \textbf{0.019} & \textbf{0.024} & 0.085 & \textbf{0.052} & \textbf{0.035} & 0.025 & \textbf{0.061} & 0.041 & \textbf{0.028} & 0.521\tnote{2} & \textbf{0.040} \\ 
& & \textit{Ours-v-nc} & \textbf{0.021} & 0.022 & 0.032 & \textbf{0.076} & 0.087 & 0.036 & \textbf{0.024} & 0.085 & \textbf{0.027} & 0.035 & 1.125\tnote{2} & 0.045 \\ \cdashline{3-15}
& & \textit{Ours-v-ba} & 0.016 & 0.024 & 0.029 & 0.074 & 0.090 & 0.035 & 0.020 & 0.060 & 0.025 & 0.024 & 1.068\tnote{2} & 0.040 \\
\cline{1-15}

\multirow{3}{*}{\rotatebox[origin=c]{90}{VIO}} & \multirow{3}{*}{\rotatebox[origin=c]{90}{Causal (C)}} &
OpenVINS~\cite{geneva2020openvins}\tnote{1} & 0.183 & 0.129 & 0.170 & \underline{0.172} & 0.212 & 0.055 & \underline{0.044} & \underline{0.069} & \underline{0.058} & \underline{0.045} & \underline{0.147} & 0.117 \\
& & Kimera2~\cite{abate2023kimera2}\tnote{1} & \underline{0.100} & \underline{0.100} & \underline{0.160} & 0.210 & \underline{0.150} & \underline{0.050} & \textbf{0.040} & 0.100 & 0.060 & 0.070 & 0.190 & \underline{0.112} \\

& & \textit{Ours-vi-c} & \textbf{0.050} & \textbf{0.048} & \textbf{0.085} & \textbf{0.144} & \textbf{0.125} & \textbf{0.045} & 0.046 & \textbf{0.046} & \textbf{0.033} & \textbf{0.030} & \textbf{0.071} & \textbf{0.066} \\ \cline{1-15}

\multirow{8}{*}{\rotatebox[origin=c]{90}{VI-SLAM}} & \multirow{2}{*}{\rotatebox[origin=c]{90}{Causal}} & \textit{Ours-vi-c} & 0.039 & \textbf{0.032} & \textbf{0.051} & 0.102 & 0.098 & \textbf{0.036} & \textbf{0.030} & \textbf{0.031} & 0.033 & \textbf{0.025} & 0.051 & 0.048 \\
& & \textit{Ours-vid-c} & \textbf{0.036} & \textbf{0.032} & 0.053 & \textbf{0.093} & \textbf{0.086} & 0.037 & 0.032 & \textbf{0.031} & \textbf{0.028} & 0.028 & \textbf{0.051} & \textbf{0.046} \\ \cline{2-15}

& \multirow{6}{*}{\rotatebox[origin=c]{90}{Non-causal (NC)}} & ORB-SLAM3~\cite{campos2021orb}\tnote{1} & 0.036 & 0.033 & 0.035 & \textbf{0.051} & 0.082 & 0.038 & \textbf{0.014} & 0.024 & 0.032 & \textbf{0.014} & 0.024 & 0.035 \\
& & MAVIS-SLAM~\cite{wang2024mavis}\tnote{1} & 0.024 & 0.025 & 0.032 & \underline{0.053} & 0.075 & \textbf{0.034} & 0.016 & 0.021 & 0.031 & 0.021 & 0.039 & 0.034 \\

& & \textit{Ours-vi-nc} & \underline{0.023} & \textbf{0.021} & \textbf{0.030} & 0.067 & \textbf{0.056} & 0.035 & \textbf{0.014} & \underline{0.020} & \underline{0.019} & 0.017 & \underline{0.022} & \textbf{0.030} \\
& & \textit{Ours-vid-nc} & \textbf{0.022} & \underline{0.023} & \textbf{0.030} & 0.068 & \underline{0.058} & \textbf{0.034} & \textbf{0.014} & \textbf{0.019} & \textbf{0.018} & \underline{0.016} & \textbf{0.020} & \textbf{0.030} \\ \cdashline{3-15}

& & \textit{Ours-vi-ba} & 0.017 & 0.024 & 0.029 & 0.057 & 0.055 & 0.035 & 0.013 & 0.020 & 0.023 & 0.013 & 0.020 & 0.028 \\
& & \textit{Ours-vid-ba} & 0.016 & 0.024 & 0.029 & 0.062 & 0.053 & 0.035 & 0.013 & 0.019 & 0.020 & 0.012 & 0.022 & 0.028 \\
\Xhline{3\arrayrulewidth}

\end{tabular}
\begin{tablenotes}
   \item All \textit{Ours} report median in 10 runs.
   \item [1] Results taken from respective papers, and OpenVINS from \cite{wang2024mavis}.
   \item [2] We consider this as a failed sequence, thus exclude this in the average.
\end{tablenotes}
\end{threeparttable}
\end{center}
\end{table*}

\subsection{Evaluation metrics}
We evaluate several variations of our method to clearly show the effectiveness of multi-modality depending on the sensor configuration as well as the estimator causality. On the one hand, in Table \ref{tab:definition_variations}, \textit{Visual} includes the reprojection (\ref{eq:residuals_reprojection}) and relative posegraph residuals (\ref{eq:residuals_rel_pose}), \textit{Inertial} indicates preintegration residuals (\ref{eq:residuals_imu}), \textit{Depth-network} is based on the depth network fusion for the submap alignment residuals (\ref{eq:method_lidar_residual}), \textit{LiDAR} uses LiDAR point clouds for the submap alignment residuals (\ref{eq:method_lidar_residual}), and \textit{GNSS} includes GNSS residuals (\ref{eq:gps res}). On the other hand, \textit{Causal} only takes measurements up to the current time, \textit{Non-causal} is the final loop-closed trajectory, and \textit{Full-BA} refines the \textit{Non-causal} trajectory in the entire states by turning relative pose graph edges to reprojection edges. For instance, \textit{Ours-vi-c} means a visual-inertial configuration with the causal evaluation.

To quantify the accuracy of the estimated poses, the estimated trajectory is aligned in $SE(3)$ to the ground-truth before evaluation. We report trajectory accuracy as root mean square error (RMSE) of the absolute trajectory error (ATE) in EuRoC and VBR datasets. 
In the Hilti-Oxford dataset, we report the predefined score, where the position error $[1,10]\,\text{cm}$ is scaled to $[0,100]$ points. To evaluate mapping accuracy, submap meshes are reconstructed using marching cubes~\cite{lorensen1998marching} and combined using submap poses. Estimated and ground-truth vertices are downsampled with a voxel size of $1\text{cm}^3$, then we perform a point-to-plane ICP alignment between both sets of vertices. We report accuracy as an average distance from all estimated vertices to the ground-truth within $0.2\,\text{m}$. 
The completeness is defined as a fraction of ground-truth vertices that are within $0.2\,\text{m}$ to the estimated vertices.

\subsection{EuRoC dataset}
This dataset provides stereo images and IMU measurements recorded by a drone, the ground-truth trajectory and mm-level accurate point clouds of the Vicon room~\cite{Euroc}. We use the stereo pair for our stereo network and left images for our MVS network. Table \ref{tab:euroc_ate} summarizes ATE where all competitors also use a stereo or stereo-inertial configuration. We additionally implemented V-SLAM, \textit{Ours-v}, to show versatility of OKVIS2-X where the IMU preintegration term is replaced by a constant velocity model. OKVIS2-X shows competitive trajectory accuracy when compared to ORB-SLAM3. It is worth noting that the \texttt{V203} sequence is challenging for V-SLAM due to the motion blur, image dropouts, and significantly varying exposure.  This is the reason why OKVIS2-X and ORB-SLAM3 present a large trajectory error on that specific sequence. 

However, we resolve this large error in a stereo-inertial configuration.
We compare the VIO implementation of \textit{Ours-vi} without loop-closure for fair comparison to odometry approaches, OpenVINS~\cite{geneva2020openvins} and Kimera2~\cite{abate2023kimera2}. Our method decreases the localization error by $41\%$ when compared to competitors. In the SLAM implementation with loop-closure, \textit{Ours-vi} outperforms state-of-the-art methods in the non-causal evaluation. It is worth noting that incorporating loop-closures yields more accurate live trajectories, reducing $1.8\,\textrm{cm}$ on average when compared to our VIO variant. Furthermore, our method with the depth networks, \textit{Ours-vid}, improves the baseline, \textit{Ours-vi}, by adding submap alignment factors in the causal evaluation. We obtained the same average accuracy when compared to the baseline in the non-causal evaluation. This indicates that the depth fusion could not achieve $3\,\text{cm}$-level accuracy ($0.04\%$ of the trajectory length), which is already quite accurate.

\begin{figure}
\centerline{\includegraphics[width=.98\linewidth]{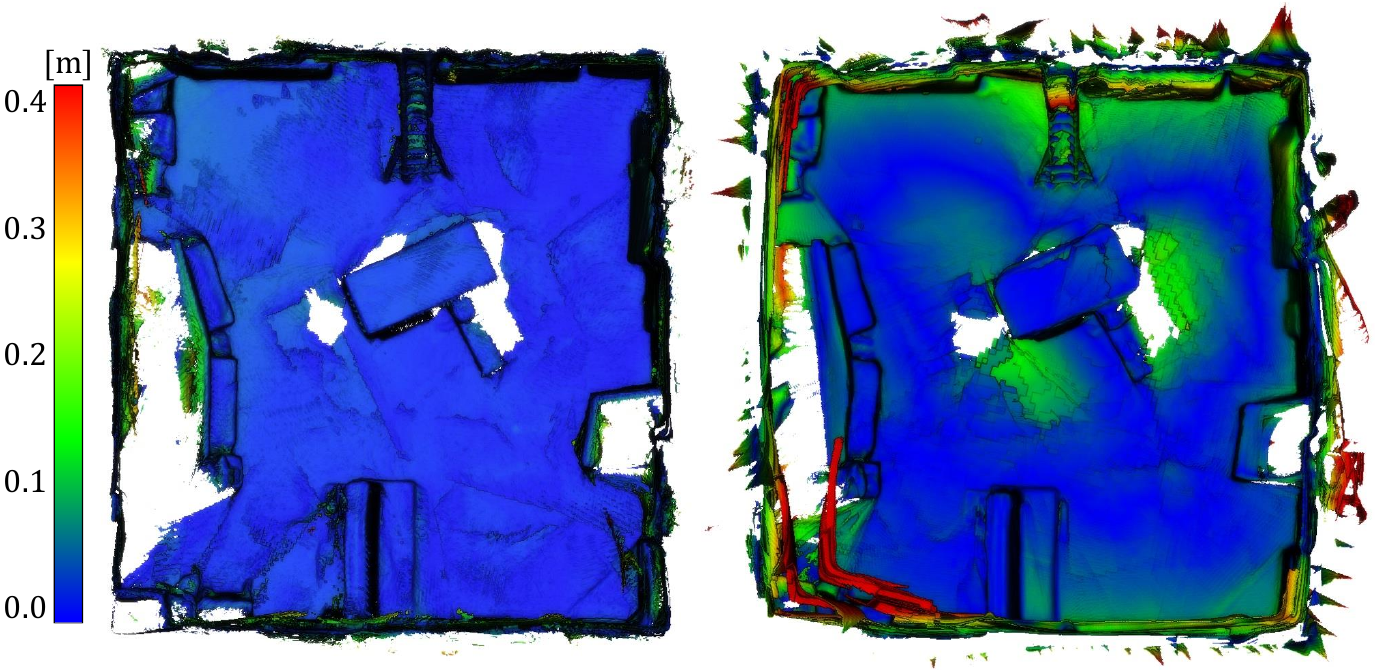}}
\caption{Reconstructed mesh with mesh-to-point error encoding in \texttt{V102} sequence. (Left) \textit{Ours-vid} and (Right) Simplemapping~\cite{xin2023simplemapping}.} \label{fig:euroc_mesh}
\end{figure}

Since our method tightly couples localization and volumetric mapping, we also evaluate the mapping accuracy in Table~\ref{tab:euroc_mesh}. To filter out noisy surfaces, we only mesh surfaces with gradients larger than a threshold and neighbors that have been observed at least three times. 
To ensure a fair comparison with SimpleMapping~\mbox{\cite{xin2023simplemapping}} in a monocular-inertial setup, we implement \textit{Ours-vid (Mono)}, where only the left camera images are used as input to the estimator and the MVS network.
Also, we present \textit{Ours-vid (Stereo)} where only depths from the stereo network are integrated for mapping. \textit{Ours-vid} achieves the highest accuracy and completeness, while \textit{Ours-vid (Mono)} also outperforms the competitor. We can clearly see the depth fusion of temporal and stereo images improves the mapping quality in Table~\ref{tab:euroc_mesh}. Fig. \ref{fig:euroc_mesh} compares the accuracy where \textit{Ours-vid} improves mesh accuracy especially along the edge of the room by properly addressing the uncertainty.

\begin{table}
\caption{Mesh accuracy [m] and completeness [\%] with 0.2m threshold in the EuRoC dataset} \label{tab:euroc_mesh}
\centering
\renewcommand{\arraystretch}{1.15} 
\begin{threeparttable}[t]
\begin{tabular}{c|c c c c|c c c c}
     & \rotatebox[origin=c]{90}{Simplemapping\tnote{1}~\cite{xin2023simplemapping}} & \rotatebox[origin=c]{90}{\textit{Ours-vid (Mono)}} & \rotatebox[origin=c]{90}{\textit{Ours-vid (Stereo)}} & \rotatebox[origin=c]{90}{\textit{Ours-vid}} & \rotatebox[origin=c]{90}{Simplemapping\tnote{1}~\cite{xin2023simplemapping}} & \rotatebox[origin=c]{90}{\textit{Ours-vid (Mono)}} & \rotatebox[origin=c]{90}{\textit{Ours-vid (Stereo)}} & \rotatebox[origin=c]{90}{\textit{Ours-vid}} \\ \\[-1em]
    
    \hline
    & \multicolumn{4}{c|}{\textit{Accuracy}} & \multicolumn{4}{c}{\textit{Completeness}} \\

    \hline
    \texttt{V101} & 0.071 & 0.050 & \underline{0.039} & \textbf{0.032} & 40.58 & \underline{43.98} & 42.70 & \textbf{44.77} \\
    \texttt{V102} & 0.077 & 0.043 & \underline{0.035} & \textbf{0.031} & 55.07 & \underline{56.45} & 56.17 & \textbf{58.38} \\
    \texttt{V103} & 0.086 & 0.062 & \underline{0.040} & \textbf{0.039} & 47.74 & 61.65 & \underline{64.59} & \textbf{67.77} \\
    
    \hline
    \texttt{V201} & \underline{0.062} & 0.066 & 0.068 & \textbf{0.047} & \underline{41.93} & 40.51 & 38.53 & \textbf{43.82} \\
    \texttt{V202} & 0.065 & \underline{0.047} & 0.064 & \textbf{0.041} & 56.37 & \underline{58.16} & 48.90 & \textbf{60.01} \\
    \texttt{V203} & 0.068 & \underline{0.062} & 0.068 & \textbf{0.044} & \textbf{62.58} & 60.20 & 52.72 & \underline{61.76} \\
    
    \hline
    Avg & 0.072 & 0.055 & \underline{0.052} & \textbf{0.039} & 50.71 & \underline{53.49} & 50.60 & \textbf{56.09} \\

    \hline
\end{tabular}
\begin{tablenotes}
   \item [1] Results obtained ourselves with the same metric.
 \end{tablenotes}
\end{threeparttable}
\end{table}

\setlength{\tabcolsep}{2.5pt}
\begin{table*}[tb]
\begin{center}
\begin{threeparttable}[b]
\caption{Localization scores in the Hilti-Oxford dataset}
\label{tab:hilti22}
\renewcommand{\arraystretch}{1.3} 
\begin{tabular}{c | c | l | c  c  c  c  c  c  c  c | c }
\Xhline{3\arrayrulewidth}
\multicolumn{2}{c|}{} & \multirow{2}{*}{\textbf{Methods}} & \multicolumn{8}{c|}{\textbf{Sequence}} & \multirow{2}{*}{\textbf{Total}} \\ \cline{4-11}
\multicolumn{2}{c|}{} & & \texttt{exp01} & \texttt{exp02} & \texttt{exp03} & \texttt{exp07} & \texttt{exp09} & \texttt{exp11} & \texttt{exp15} & \texttt{exp21} \\
\Xhline{3\arrayrulewidth}

\multirow{10}{*}{\rotatebox[origin=c]{90}{Visual-inertial}} & \multirow{4}{*}{\rotatebox[origin=c]{90}{Causal}} & OpenVINS~\cite{geneva2020openvins}\tnote{1} & 0.00 & 0.00 & 0.00 & 1.67 & 8.75 & 8.00 & 4.44 & 0.00 & 22.86 \\

& & \multirow{2}{*}{\textit{Ours-vi-c}} & 10.77 & 8.18  & 0.00 & 0.00 & 11.25 & 10.00 & 10.00 & 0.00 & 50.20 \\[-0.5em]
& & & (7.69) & (10.00) & (0.00) & (0.00) & (6.25) & (6.00) & (11.11) & (0.00) & (41.05) \\

& & \multirow{2}{*}{\textit{Ours-vid-c}} & \textbf{56.92} & 18.18 & 0.00 & 0.00 & 4.38 & 4.00 & 7.78 & 0.00 & 91.26 \\[-0.5em]
& & & (26.15) & (4.55) & (0.00) & (1.67) & (1.88) & (8.00) & (4.44) & (0.00) & (46.69) \\ \cline{2-12}

& \multirow{6}{*}{\rotatebox[origin=c]{90}{Non-causal}} & BAMF-SLAM~\cite{zhang2023bamf}\tnote{2} & 15.38 & 4.55 & 8.24 & 1.67 & 1.25 & 2.00 & 7.78 & 0.00 & 40.86 \\

& & ORB-SLAM3~\cite{campos2021orb}\tnote{1} & 3.85 & 1.82 & 10.59 & 0.00 & 0.00 & 6.00 & 0.00 & 0.00 & 22.25 \\

& & MAVIS SLAM~\cite{wang2024mavis}\tnote{2} & 40.00 & 16.82 & \textbf{16.47} & \underline{11.67} & 15.00 & 20.00 & \underline{26.67} & \textbf{4.00} & 150.62 \\

& & \multirow{2}{*}{\textit{Ours-vi-ba}} & 30.00 & \textbf{40.91} & \underline{12.35} & 3.33 & \underline{16.88} & \textbf{44.00} & 20.00 & 0.00 & \underline{167.47} \\[-0.5em]
& & & (36.92) & (30.45) & (0.00) & (1.67) & (17.50) & (28.00) & (20.00) & (0.00) & (135.54) \\

& & \multirow{2}{*}{\textit{Ours-vid-ba}} & \underline{53.08} & \underline{34.55} & 0.00 & \textbf{13.13} & \textbf{23.75} & \underline{22.00} & \textbf{28.89} & \underline{2.00} & \textbf{177.59} \\[-0.5em]
& & & (30.00) & (37.27) & (0.00) & (8.33) & (23.75) & (22.00) & (26.67) & (0.00) & (148.02) \\
\hline

\multirow{7}{*}{\rotatebox[origin=c]{90}{(V)I-LiDAR}} & \multirow{3}{*}{\rotatebox[origin=c]{90}{ Causal}} &
FAST-LIVO2~\cite{zheng2024fastlivo2}\tnote{2} & \textbf{84.62} & \underline{70.00} & 44.71 & \underline{31.67} & 7.50 & \underline{92.00} & 24.44 & 48.00 & 402.93\\
& & \multirow{2}{*}{\textit{Ours-vil-c}} & 63.08 & 33.64 & 1.18 & 0.00 & 11.88 & 68.00 & 30.00 & 38.00 & 245.76 \\[-0.5em]
& & & (50.77) & (37.27) & (7.06) & (0.00) & (7.50) & (62.00) & (34.44) & (34.00) & (233.05) \\ \cline{2-12}

& \multirow{4}{*}{\rotatebox[origin=c]{90}{ Non-causal}} & VILENS~\cite{vilens}\tnote{2} & 69.23 & 49.09 & 40.00 & 16.67 & 5.00 & 68.00 & 17.78 & 60.00 & 325.77\\
& & \multirow{2}{*}{\textit{Ours-vil-ba}} & 63.85 & 51.82 & \underline{49.41} & 1.67 & \underline{26.25} & \textbf{100.00} & \textbf{61.11} & \textbf{84.00} & \underline{438.10} \\[-0.5em]
& & & (72.31) & (51.36) & (51.18) & (0.00) & (24.38) & (84.00) & (61.11) & (68.00) & (412.33) \\ \cdashline{3-12}
& & Wildcat~\cite{Wildcat}\tnote{2,3} & \textbf{84.62} & \textbf{84.55} & \textbf{90.59} & \textbf{45.00} & \textbf{44.38} & 84.00 & \underline{46.67} & \textbf{84.00} & \textbf{563.79} \\
\Xhline{3\arrayrulewidth}

\end{tabular}
\begin{tablenotes}
   \item All \textit{Ours} report the best (median) scores in 3 runs based on the total score.
   \item Due to the nature of the challenge, best (bold) and second (underlined) scores are highlighted per sensor configuration without distinction of causal or non-causal methods.
   \item [1] Results, best scores in 3 runs, obtained from us based on the open-source in stereo-inertial setup.
   \item [2] Results taken from the leaderboard~\cite{hilti_leaderboard}.
   \item [3] Wildcat uses LiDAR-Inertial only.
\end{tablenotes}
\end{threeparttable}
\end{center}
\end{table*}

\begin{figure}[b]
    \centering
    \includegraphics[width=.9\linewidth]{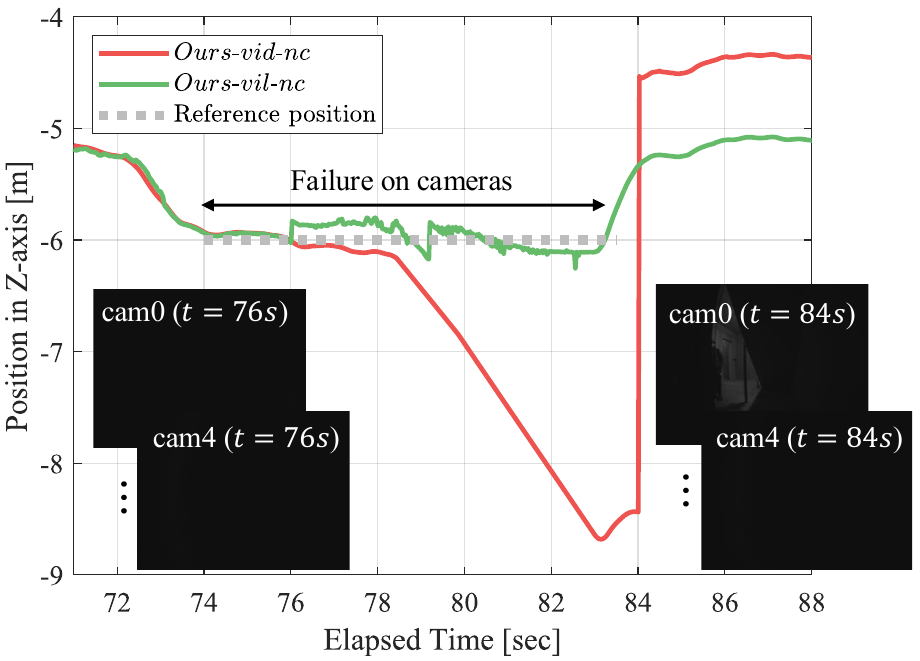}
    \caption{Failure case in \texttt{exp03} in the Hilti-Oxford dataset where the LiDAR compensates the vision failure in a dark room.}
    \label{fig:exp03-zpos}
\end{figure}

\subsection{Hilti-Oxford dataset}

This dataset provides images from 5 cameras, IMU measurements, 32-channel LiDAR point clouds captured by a handheld device, sparse ground-truth positions, and mm-accurate dense point clouds~\cite{Hilti22}. We use all 5 cameras for the visual-inertial estimator, 2 front cameras for the stereo network, and the front left camera for the MVS network. We calibrate camera-IMU extrinsic parameters online, and the close inspection of the online calibration will be presented in Sec.~\ref{sec:abl_calib}.

Table \ref{tab:hilti22} summarizes the localization scores in the challenge sequences. Considering the nature of the leaderboard, we compare best scores in 3 runs of our method to competitors. To give insight about the deviation from run-to-run, we also report the median scores in brackets. For the visual-inertial configuration, our non-causal evaluation with the final bundle adjustment significantly improves the position accuracy while outperforming published competitors in the leaderboard. By having additional residuals from submap alignment, \textit{Ours-vid-ba} further improves the accuracy. \texttt{exp03} was challenging due to a low light condition which leads temporarily to completely black images in the front cameras, as shown in Fig.~\ref{fig:exp03-zpos}. Given that we only use front cameras for dense mapping, \textit{Ours-vid-ba} records $1.2 \, \text{m}$ position error, which is relatively higher than in the other sequences, in the control points. However, the score of \textit{Ours-vid-ba} corresponds to $7.0 \, \text{cm}$ position error on average excluding the challenging sequence, \texttt{exp03}.

Our Visual-Inertial-LiDAR configuration, \textit{ours-vil-}$\star$, shows a significant improvement in both the causal and the refined estimates for almost every sequence, especially also robustifying the system in the visually challenging sequence \texttt{exp03} (Fig.~\ref{fig:exp03-zpos}). In terms of the actual localization accuracy, it reduces the average position error to $4.1 \, \text{cm}$ ($2.8 \, \text{cm}$ when excluding \texttt{exp07} which has an accuracy around $11 \, \text{cm}$) which is in the range of the mapping resolution.
We also compare our system to the top-ranked published competitors. Ours ranks top amongst the published Visual-Inertial-LiDAR systems, on average outperforming FAST-LIVO2~\cite{zheng2024fastlivo2} and VILENS~\cite{vilens} while being outperformed only by Wildcat~\cite{Wildcat}, a purely LiDAR-Inertial SLAM system.
The main bottleneck of our system is in \texttt{exp07}, a long corridor which is a degenerative case for the LiDAR due to the lack of characteristic geometry in the direction of movement.

In terms of mapping, Table~\ref{tab:hilti_mesh} reports the accuracy and completeness with the network depth and LiDAR in \texttt{exp04}, \texttt{exp05}, \texttt{exp06} where ground-truth point clouds are available. We obtain fairly accurate mapping ($4.2 \, \text{cm}$) by the depth fusion which was only fine-tuned in a synthetic dataset~\cite{wang2020tartanair}. We further improve the accuracy and completeness by replacing depth networks with a LiDAR. The \textit{far plane}, the maximum valid depth to integrate measurements, was set to $3\,\text{m}$ and $30\,\text{m}$ for the depth fusion and LiDAR, respectively. Fig.~\ref{fig:eval-meshes} visualizes reconstructed meshes in \texttt{exp06} where geometric structures captured by our depth network are clearly visible, and fusion with a LiDAR further increases the completeness.

\setlength{\tabcolsep}{6.0pt}
\begin{table}
\caption{Mesh accuracy [m] and completeness [\%] with 0.2m threshold in the HILTI dataset} \label{tab:hilti_mesh}
\centering
\renewcommand{\arraystretch}{1.3} 
\begin{tabular}{c|c c|c c}
     & \rotatebox[origin=c]{90}{\textit{Ours-vid}} & \rotatebox[origin=c]{90}{\textit{Ours-vil}}& \rotatebox[origin=c]{90}{\textit{Ours-vid}} & \rotatebox[origin=c]{90}{\textit{Ours-vil}} \\
    
    \hline
    & \multicolumn{2}{c|}{\textit{Accuracy}} & \multicolumn{2}{c}{\textit{Completeness}} \\

    \hline
    \texttt{exp04} & 0.040 & \textbf{0.030} & 59.15 & \textbf{77.90}  \\
    \texttt{exp05} & 0.040 & \textbf{0.031} & 64.04 & \textbf{76.94}  \\
    \texttt{exp06} & 0.046 & \textbf{0.027} & 69.63 & \textbf{80.15}  \\
    
    \hline
    Avg & 0.042 & \textbf{0.029} & 64.27 & \textbf{78.33}  \\

    \hline
\end{tabular}
\end{table}

\setlength{\tabcolsep}{2.5pt}
\begin{table*}[tb]
\begin{center}
\begin{threeparttable}[b]
\caption{Root mean square of absolute trajectory error [meters] in the VBR dataset.} 
\label{tab:vbr}
\renewcommand{\arraystretch}{1.4} 
\begin{tabular}{c | c | l | c  c  c  c  c  c  c  c | c }
\Xhline{3\arrayrulewidth}
\multicolumn{2}{c|}{} & \multirow{2}{*}{\textbf{Methods}} & \multicolumn{8}{c|}{\textbf{Sequence}} & \multirow{2}{*}{\textbf{Avg}} \\ \cline{4-11}

 \multicolumn{2}{c|}{} & & \texttt{Colosseum} & \texttt{Diag} & \texttt{Pincio} & \texttt{Spagna} & \texttt{Campus0} & \texttt{Campus1} & \texttt{Ciampino0} & \texttt{Ciampino1} \\
\Xhline{3\arrayrulewidth}

\multirow{7}{*}{\rotatebox[origin=c]{90}{Visual-inertial}} & \multirow{3}{*}{\rotatebox[origin=c]{90}{Causal}} & OpenVINS\cite{geneva2020openvins} & 20.827 & 24.628 & 6.928 & 10.023 & 18.343 & 4.719 & 109.777 & 34.921 & 28.770 \\
& & \textit{Ours-vi-c} & 1.922 & \textbf{1.156} & \textbf{3.368} & 3.618 & \textbf{3.065} & 2.631 & \textbf{3.721} & \textbf{2.971} & \textbf{2.807} \\

& & \textit{Ours-vid-c} & \textbf{1.771} & 1.173 & 3.673 & \textbf{3.569} & 3.084 & \textbf{2.552} & 3.728 & 3.002 & 2.819 \\ \cline{2-12}

& \multirow{5}{*}{\rotatebox[origin=c]{90}{Non-causal (NC)}} & ORB-SLAM3~\cite{campos2021orb}\tnote{1} & 7.713 & 1.834 & \textbf{1.594} & 3.108 & 8.023\tnote{2} & 7.289 & -\tnote{2} & 9.443\tnote{2} & 5.572\tnote{2}  \\

& & \textit{Ours-vi-nc} & \underline{1.541} & \underline{1.026} & 3.301 & \textbf{0.605} & \textbf{3.026} & \underline{2.499} & \underline{3.490} & \textbf{2.847} & \underline{2.292} \\

& & \textit{Ours-vid-nc} & \textbf{1.450} & \textbf{0.452} & \underline{3.150} & \underline{0.744} & \underline{3.037} & \textbf{2.482} & \textbf{3.472} & \underline{2.889} & \textbf{2.210} \\ \cdashline{3-12}

& & \textit{Ours-vi-ba} & 1.613 & 0.931 & 3.278 & 0.662 & 3.017 & 2.471 & 3.358 & 2.920 & 2.281 \\

& & \textit{Ours-vid-ba} & 1.441 & 0.346 & 3.199 & 0.807 & 3.016 & 2.451 & 3.391 & 2.995 & 2.206  \\
\hline

\multirow{4}{*}{\rotatebox[origin=c]{90}{VI-LiDAR}} & \multirow{2}{*}{\rotatebox[origin=c]{90}{Causal}} & FAST-LIVO~\cite{zheng2022fast}\tnote{1} & 6.459 & 1.013 & \textbf{2.042} & 3.963 & \textbf{1.259} & \textbf{0.434} & 3.080 & \textbf{1.548} & 2.475 \\
& & \textit{Ours-vil-c} & \textbf{1.517} & \textbf{0.900} & 2.861 & \textbf{0.691} & 1.955 & 1.634 & \textbf{1.716} & 2.892 & \textbf{1.771} \\ \cline{2-12}

& \multirow{2}{*}{\rotatebox[origin=c]{90}{NC}} & \textit{Ours-vil-nc} & 0.275 & 0.906 & 1.613 & 0.326 & 1.529 & 0.868 & 0.529 & 0.582 &   0.829  \\ \cdashline{3-12}
& & \textit{Ours-vil-ba} & 0.337 & 0.891 & 1.745 & 0.345 & 1.495 & 0.875 & 0.502  & 0.606 & 0.850  \\
\Xhline{3\arrayrulewidth}

\multicolumn{3}{c|}{Trajectory length [km]} & 1.451  & 1.037 & 1.277 & 1.562 & 2.735 & 2.925 & 9.008 & 5.200 & 3.149 \\ 
\Xhline{3\arrayrulewidth}

\end{tabular}
\begin{tablenotes}
   \item All methods report median in 3 runs.
   \item [1] Results obtained from us based on the open-source implementation.
   \item [2] System failure: Reported numbers are obtained from less than 3 successful runs (out of min. 5 runs) if any available.
\end{tablenotes}
\end{threeparttable}
\end{center}
\end{table*}

\subsection{VBR: A Vision Benchmark in Rome}
\label{sec:vbr}
The VBR dataset \cite{VBR} presents an extensive sensor suite that allows us to evaluate OKVIS2-X with different configurations. We consider VBR an interesting dataset because it contains long sequences, proving the scalability of OKVIS2-X. The recorded sequences are in real-world environments, posing additional challenges such as: dynamic entities, non-constant illumination conditions, structured and unstructured environments. This dataset contains images from a stereo camera, IMU measurements, LiDAR point clouds and ground-truth poses for quantitative evaluation. We found out that the MVS network outputs noisy depths in the open-sky due to the depth plane parametrisation which is not in inverse depth. In VBR, specifically, we only employ the stereo network. To further improve the stereo disparity accuracy, we finetuned our uncertainty-augmented network in a 2-level image pyramid followed by the additional refinement step~\cite{xu2023unifying}. In this dataset, we evaluate the ATE of OKVIS2-X in all its multi-sensor configurations and compare it to the state-of-the-art systems ORB-SLAM3~\cite{campos2021orb}, OpenVINS\cite{geneva2020openvins} and FAST-LIVO~\cite{zheng2022fast}. To ensure that FAST-LIVO processes all measurements, the rosbags were played at a real-time rate of $0.2.$

Table \ref{tab:vbr} presents the trajectory accuracy of the evaluated SLAM systems in the VBR dataset, all of them tested with the same calibration parameters. The table presents the median trajectory error from the three evaluation runs. Since VBR presents very long sequences, up to $9\,\text{km}$, we set the submap resolution to either $10$ or $20\,\text{cm}$, depending on the sensor configuration and scale of the sequence. The sensor far plane is limited to $30\,\text{m}$ for the LiDAR and $10$ or $15\,\text{m}$ for the stereo network in the handheld or driving sequences.

\begin{figure*}[t]
    \centering
    \includegraphics[width = \textwidth]{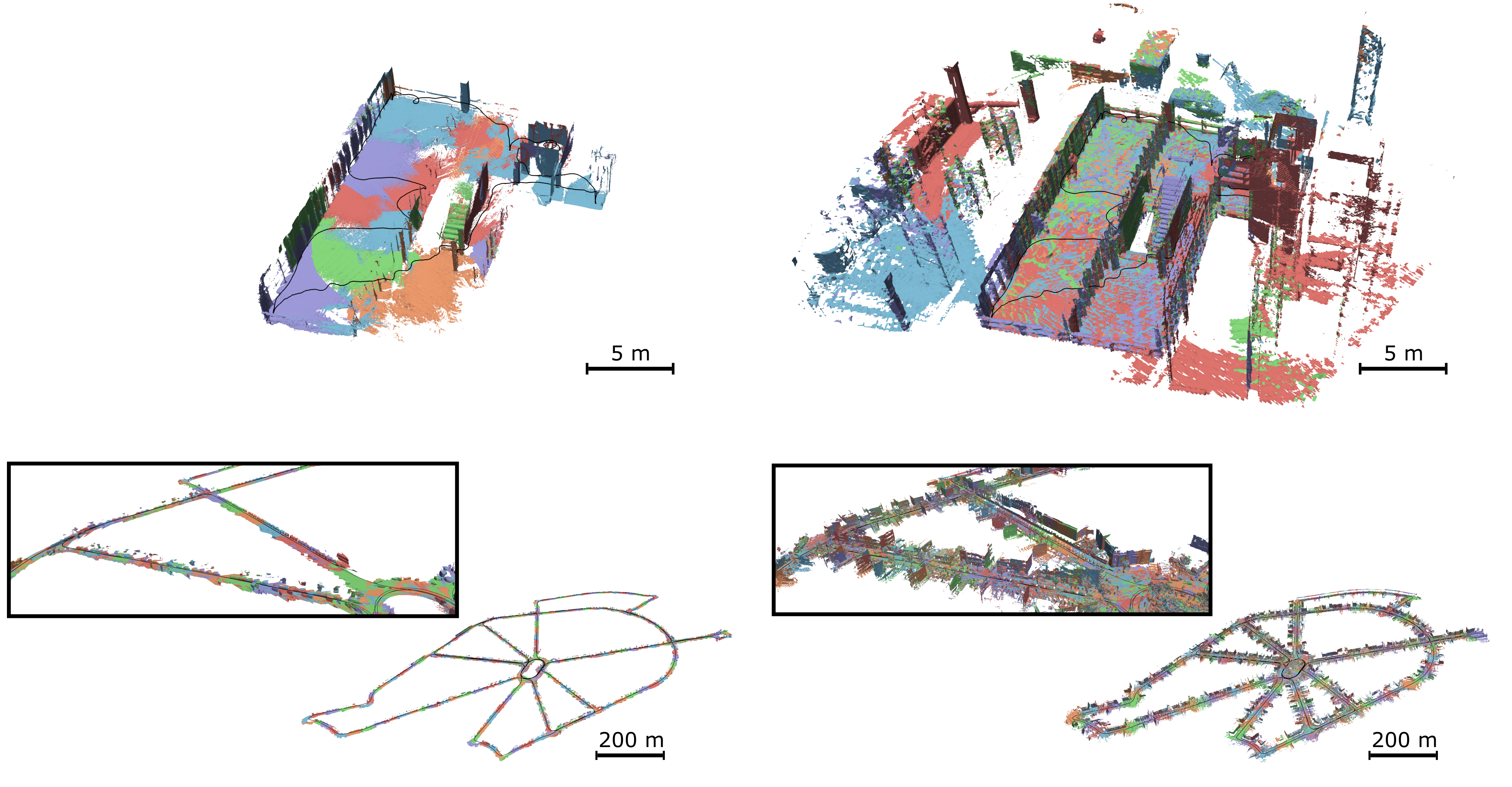}
    \caption{3D reconstructions obtained from the volumetric submaps of OKVIS2-X. The left column showcases reconstructions when leveraging learned-depth while the ones on the right use a LiDAR sensor. The top row is obtained from the Hilti22 \texttt{exp06} sequence, the bottom row is extracted from VBR's \texttt{Ciampino0} sequence. A line with a distance indicates the different scales of the sequences.}
    \label{fig:eval-meshes}
\end{figure*}

In the visual-inertial sensor configuration OKVIS2-X demonstrates a superior performance in most sequences, with the version that leverages learned depth being the most accurate in the non-causal evaluation. Since OpenVINS does not support loop-closures, its accuracy naturally degrades in large-scale scenarios with loops. In the VI-LIDAR sensor configuration, OKVIS2-X demonstrates an overall superior performance over FAST-LIVO already in the causal evaluation, with an average error of $1.771\,\textrm{m}$ ($0.06\%$ of the trajectory). Our system's support for loop-closures leads to another significant improvement. Figures~\ref{fig:top-figure} and~\ref{fig:eval-meshes} show examples of globally consistent submap meshes.

OKVIS2-X demonstrates a superior performance over the compared systems. Furthermore, OKVIS2-X showcases its robustness in the presence of dynamic entities, especially pronounced in \texttt{Colosseum} or \texttt{Spagna}, or in case of corrupted data, as we found that the driving sequences \texttt{Campus*} and \texttt{Ciampino*} contain episodes of missing IMU measurements throughout the sequence. In these cases, the other systems suffered from degrading accuracy or system failure.

\begin{figure*}[bt]
    \centering
    \includegraphics[width=.90\linewidth]{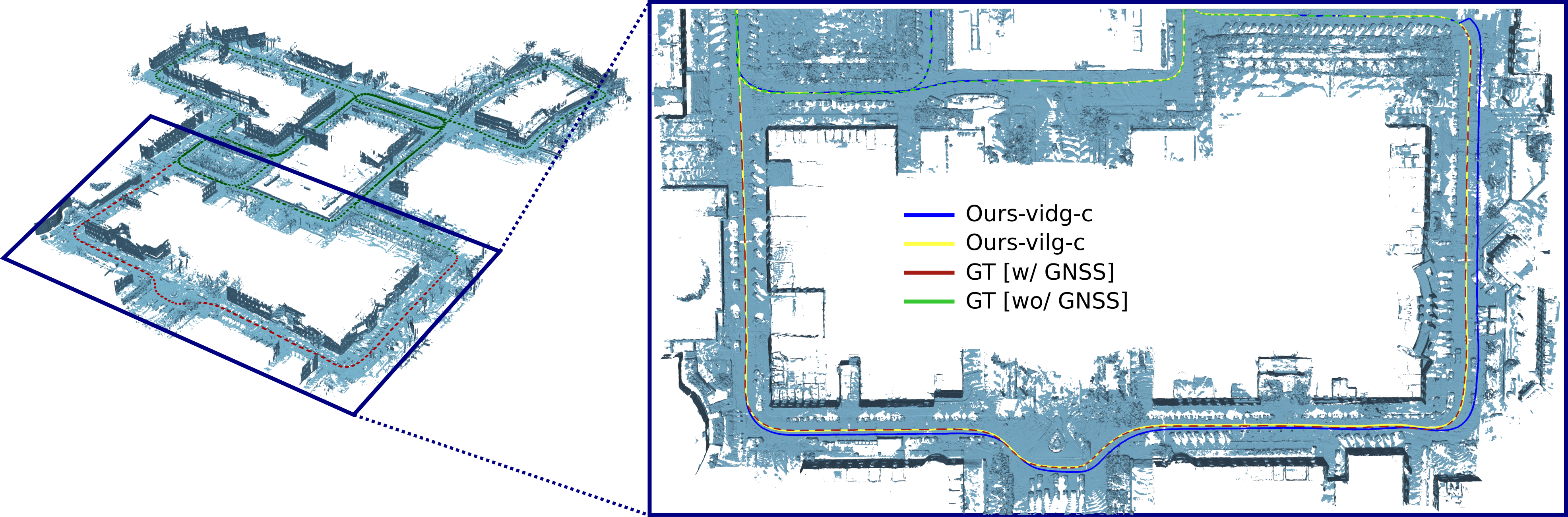}
    \caption{Reconstruction of the \texttt{Campus1} (VBR) is visualized as overlay of all submaps. Causal Trajectory Estimates for our Visual-Inertial-Depth (blue) and Visual-Inertial-LiDAR (yellow) configurations with GNSS fusion are drawn together with ground-truth measurements in GNSS-accessible (dashed green) and GNSS-denied regions (dashed red). While the LiDAR-based setup aligns accurately with the ground-truth even in GNSS-denied areas, the vision-based configuration drifts significantly. The visible jump in the trajectory at the end of the GNSS dropout highlights the global alignment strategy of our approach.}
    \label{fig:eval-gnss}
\end{figure*}

\setlength{\tabcolsep}{4.0pt}
\begin{table}
\caption{Dropout-Tolerant GNSS Fusion: ATE RMSE[m] in VBR \texttt{Campus1}.} \label{tab:gnss_vbr}
\centering
\renewcommand{\arraystretch}{1.3} 
\begin{tabular}{c| c c | c c | c c}
     & \rotatebox[origin=c]{90}{\textit{Ours-vi}} & \rotatebox[origin=c]{90}{\textit{Ours-vig}} & \rotatebox[origin=c]{90}{\textit{Ours-vid}} & \rotatebox[origin=c]{90}{\textit{Ours-vidg}} & \rotatebox[origin=c]{90}{\textit{Ours-vil}} & \rotatebox[origin=c]{90}{\textit{Ours-vilg}} \\
    
    \hline
    Causal & 2.631 & 1.119 & 2.552 & \underline{1.099} & 1.634 & \textbf{0.328}  \\
    (during Dropout) & - & (2.766) & - & (2.737) & - & (0.701) \\
    \cdashline{1-7}
    Non-Causal & 2.499 & 0.751 & 2.482 & \underline{0.750} & 0.868 & \textbf{0.177}  \\
    (during Dropout) & - & (1.974) & - & (1.972) & - & (0.423) \\
    \cdashline{1-7}
    Final BA & 2.471 & 0.661 & 2.451 & \underline{0.636} &  0.875 & \textbf{0.169} \\  
    (during Dropout) & - & (1.719) & - & (1.655) & - & (0.386) \\
    \hline
\end{tabular}

\begin{tablenotes}
   \item All methods report median in 3 runs.
\end{tablenotes}
\end{table}

\subsection{Dropout-Tolerant Fusion of Global Position Measurements}
\label{sec:eval-gnss}
To demonstrate the versatility and modularity of OKVIS2-X for different sensor setups, and to demonstrate the capability to fuse global position measurements, we conducted two more studies. The first one is on the VBR sequence \texttt{Campus1}, one of the driving sequences. As the dataset does not provide geodetic GNSS measurements, we simulated RTK-GNSS measurements by applying Gaussian noise of $1\,\text{cm}$ in horizontal and $2\,\text{cm}$ in vertical direction to the ground-truth trajectory. Furthermore, we simulated a GNSS dropout of 75 seconds and a corresponding trajectory length of $450\,\text{m}$. To mimic a realistic scenario, the dropout was chosen in an area of the sequence that has more narrow streets and higher buildings than on average. The main target of this evaluation is to give a proof-of-concept for the fusion of global position measurements and especially to showcase the effectiveness of the global, loop-closure-like, alignment strategy to handle potential signal outages over longer periods of time, presented in Sec.~\ref{sec:SLAM-gnss-align}.
Table~\ref{tab:gnss_vbr} reports the ATE for all possible sensor setups (\textit{vig}, \textit{vidg} and \textit{vilg}) for the whole trajectory as well as during the GNSS signal outage only.
Obviously, the overall trajectory error is significantly reduced with the fusion of GNSS measurements in all configurations. Furthermore, it can be seen that the accuracy of our Visual-Inertial-LiDAR configuration helps to minimize the drift during the GNSS dropout. Loop-closures and a final BA enable us to estimate a final trajectory with an RMSE ATE of only  $16.9\,\text{cm}$ for a trajectory of almost $3\,\text{km}$.
Qualitative results on the reconstruction and trajectory accuracy are visualized in Fig.~\ref{fig:eval-gnss}. 

\begin{figure}[t]
    \centering
    \includegraphics[width=.9\linewidth]{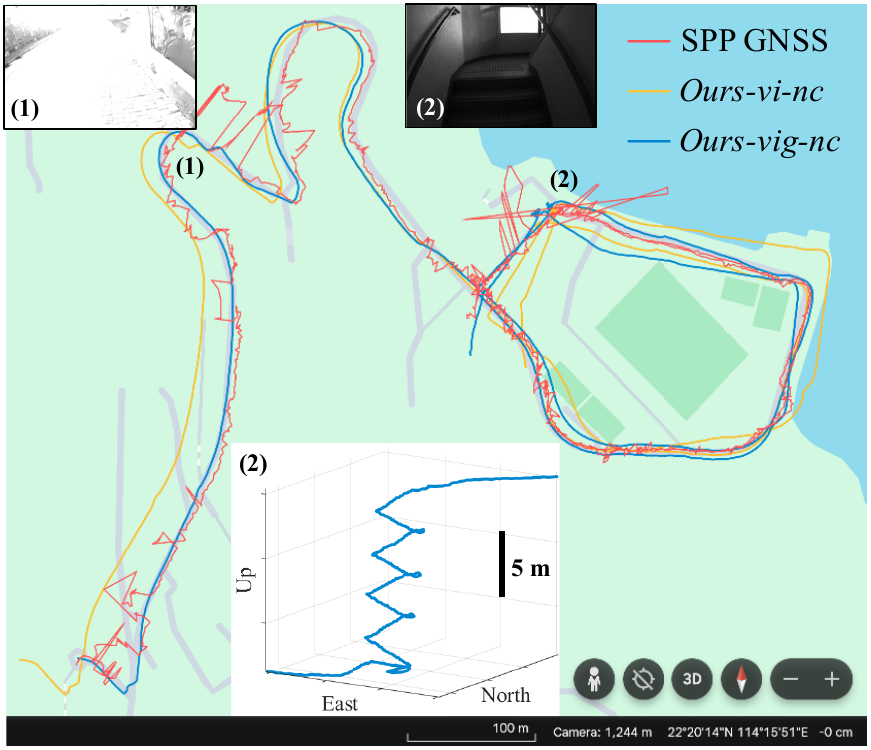}
    \caption{SPP Solution of GNSS and OKVIS2-X estimated trajectories overlaid in Google Maps in the \texttt{complex}\_\texttt{environment} in the GVINS dataset.}
    \label{fig:complex-gvins}
\end{figure}

Even though the system was originally designed to fuse intermittent high-accuracy (RTK-) GNSS measurements, it can still also benefit from lower-grade GNSS measurements. To demonstrate that, we evaluated OKVIS2-X on one of the challenging dataset sequences provided by GVINS\mbox{\cite{GVINS}}. This dataset consists of a stereo camera, an IMU and a ZED-F9P GNSS sensor. The authors provide RTK solutions as well as raw GNSS measurements (e.g.\ pseudoranges, Doppler measurements, etc.). We chose to evaluate on the nearly $3$ km \texttt{complex\_environment} sequence due to its numerous challenges for different sensor modalities: featureless images in very bright and dark areas as well as high corruption or complete unavailability of GNSS signals in cluttered environments or indoors (see Fig.~\mbox{\ref{fig:complex-gvins}}). Using RTKLIB~\mbox{\cite{rtklib}}, we computed the Single Point Positioning (SPP) solution from raw measurements and fused it into the factor graph optimization. We evaluate the RMSE ATE after 6DoF alignment with the RTK solution as a reference in areas with a valid RTK fix. Even with the highly imprecise SPP solution (RMSE ATE of $19.99\,\textrm{m}$), the proposed GNSS fusion significantly reduced the error of the visual-inertial baseline from $23.72$\,\textrm{m} to $12.51$ (causal), $8.47$ (non-causal) and $5.12\,\textrm{m}$ (final BA). Note that in this evaluation, visual loop closures have been disabled for a fairer comparison with GVINS\mbox{\cite{GVINS}}, which achieves an RMSE ATE of $4.45\,\textrm{m}$. This shows that, despite the simpler fusion strategy of GNSS measurements with less precise data (SPP only, and no Doppler velocities), the overall accuracy is still competitive. Fig.~\mbox{\ref{fig:complex-gvins}} shows the estimated trajectories overlaid with the SPP solution. It is clearly visible that, compared to the visual-inertial baseline, the fused estimate aligns well with a road layout in areas of good coverage but stays consistent even in areas of highly corrupted GNSS signals.

\begin{figure*}[t]
    \centering
    \includegraphics[width = \linewidth]{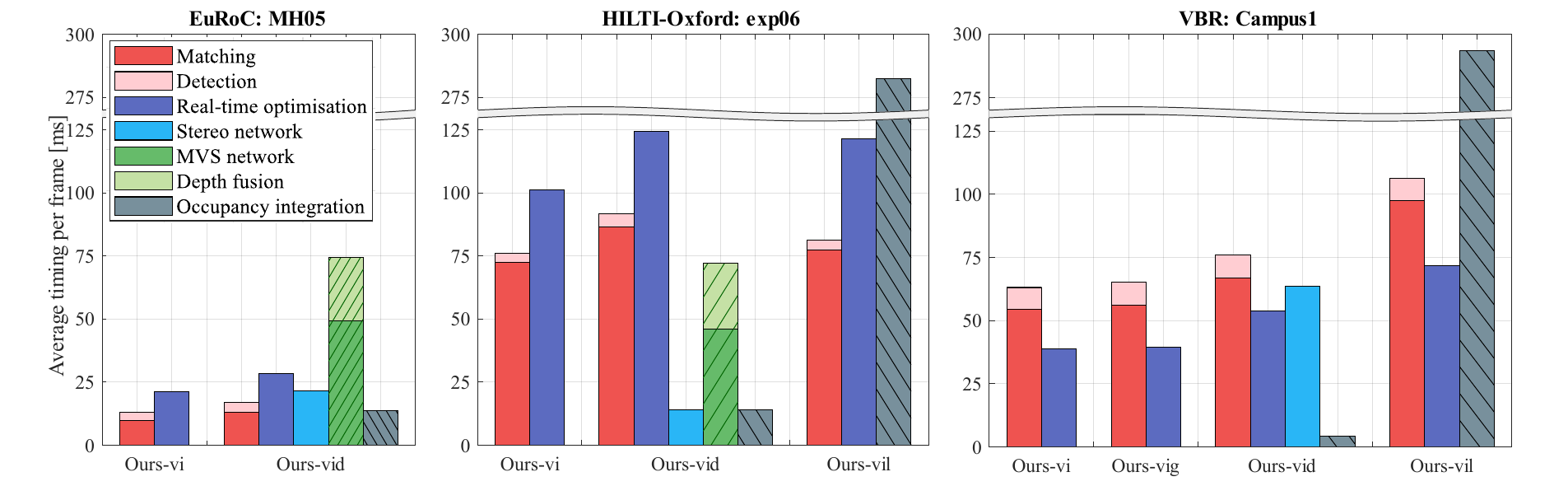}
    \caption{Timing breakdown of OKVIS2-X with different configurations in ms per frame. Each bar represents a thread, and multiple threads run in parallel, therefore, the total timing is not the sum of each bar. The networks, depth fusion, and map integration do not necessarily have to be performed for every frame or LiDAR scan. Also, HILTI-Oxford uses five cameras on the rig, and VBR employs high-resolution cameras amenable to higher keypoint numbers.}
    \label{fig:timings}
\end{figure*}

\setlength{\tabcolsep}{3.0pt}
\begin{table}
\caption{Ablation study in online extrinsic calibration: ATE RMSE [m] in the Hilti-Oxford dataset} \label{tab:abl_hilti}
\centering
\begin{threeparttable}[b]
\renewcommand{\arraystretch}{1.1} 
\begin{tabular}{l | c c | c c c | c}
    & \textit{Online}- & \textit{Post}- & \multirow{2}{*}{\texttt{exp04}} & \multirow{2}{*}{\texttt{exp05}} & \multirow{2}{*}{\texttt{exp06}} & \multirow{2}{*}{Avg} \\[-2pt]
    & \textit{calibration} & \textit{calibration} & & & & \\
    \hline
    
    \multirow{3}{*}{\textit{Ours-vi-c}} & X & X & 0.198 & 0.192 & 0.085 & 0.158 \\
    & \checkmark & X & 0.070 & 0.058 & 0.067 & 0.065 \\
    & X & \checkmark & \textbf{0.058} & \textbf{0.044} & \textbf{0.048} & \textbf{0.050} \\
    \hline

    \multirow{3}{*}{\textit{Ours-vi-nc}} & X & X & 0.157 & 0.169 & 0.090 & 0.139 \\
    & \checkmark & X & 0.036 & 0.049 & \textbf{0.041} & 0.042 \\
    & X & \checkmark & \textbf{0.033} & \textbf{0.026} & 0.042 & \textbf{0.034} \\
    \hline

    \multirow{3}{*}{\textit{Ours-vi-ba}} & X & X & 0.096 & 0.088 & 0.061 & 0.082 \\
    & \checkmark & X & 0.023 & \textbf{0.022} & 0.050 & \textbf{0.032} \\
    & X & \checkmark & \textbf{0.020} & 0.028 & \textbf{0.048} & \textbf{0.032} \\
    \hline
\end{tabular}
\begin{tablenotes}
   \item All methods report median in 3 runs.
\end{tablenotes}
\end{threeparttable}
\end{table}

\subsection{Ablation study: Online extrinsic calibration}
\label{sec:abl_calib}
To show effectiveness of the IMU-to-camera extrinsic online calibration, Table \ref{tab:abl_hilti} reports an ablation study of how online calibration affects the estimated trajectory accuracy where \textit{Online-calibration} indicates an online optimization of the extrinsic parameters, and \textit{Post-calibration} means that the extrinsic parameters are initialized with our calibrated extrinsics and kept constant. We chose \texttt{exp04}, \texttt{exp05}, and \texttt{exp06} sequences of the Hilti-Oxford dataset where the dense ground-truth trajectory is available. In the live trajectory, we set $1\text{mm}$ standard deviation for the prior translation and $0.29^\circ$ for the prior orientation for all five cameras. In the full bundle adjustment, where well-initialized poses and landmarks are available, we apply weaker constraints on the standard deviation to give more freedom in optimizing extrinsics: $3\text{mm}$ and $0.92^\circ$. 
It can be clearly seen that the online calibration improves the trajectory accuracy in all sequences in Table \ref{tab:abl_hilti} when compared to the case without calibration. To show the validity of the optimized extrinsic parameters, we evaluate additional runs with our extrinsic parameters without online calibration. Interestingly, our \textit{Post-calibration} improves the online calibration results. This indicates that our estimated extrinsics are close to the ground-truth extrinsics. Furthermore, to show the robustness to the initial error, we perturbed the initial extrinsic parameters from our extrinsics after the full-BA calibration. with an uniform distribution of $\pm 1\,\textrm{cm}$ and $\pm 0.3^\circ$ in \texttt{exp06}. Considering 10 runs, the extrinsics record a standard deviation of only $0.08\,\mathrm{cm}$ and $0.02^\circ$ at the end of the sequence. This indicates successful convergence of the perturbed extrinsics to the reference in our online calibration.

\begin{figure}
\centerline{\includegraphics[width=\linewidth]{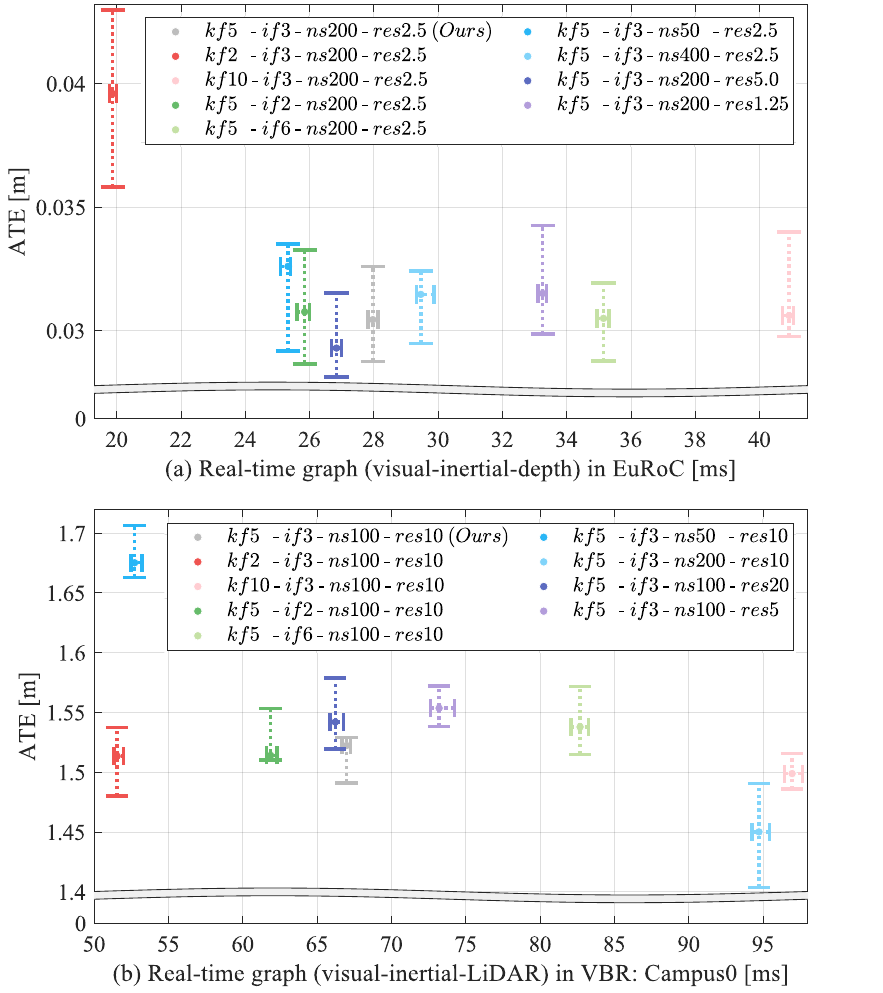}}
\caption{Parameter study on OKVIS2-X (a) across all sequences in EuRoC and (b) \texttt{Campus0} in the VBR dataset with the number of keyframes $kf$, imu frames $if$, submap factors $ns$, and the voxel resolution $res$ in cm. Each vertical bar represents 10, 50, 90 precentiles in non-causal ATE, while the horizontal bar means 10, 50, 90 percentiles in the real-time optimization timings per frame. The statistics comes from 10 independent runs.} \label{fig:parameter}
\end{figure}

\subsection{Configurable parameter study}
We thoroughly investigate trajectory errors and real-time graph optimization timings of $vid$ and $vil$ modes with respect to the number of keyframes ($kf$), imu frames ($if$), submap factors ($ns$), and the voxel resolution ($res$), as shown in Fig.~\ref{fig:parameter}. For the submap-based alignment factors, we set the number of map-to-map factors to a multiple of the number of frame-to-map factors (i.e.\ $5ns$).

For the $vid$ configuration in Fig.~\ref{fig:parameter}(a), increasing the number of keyframes and IMU frames results in longer processing times, while the accuracy is saturated at the proposed parameter configuration
(gray bars). Similarly, adding more submap factors and using finer voxel resolutions increases computation time but does not necessarily improve accuracy, as the estimated trajectory already exceeds the accuracy of the network depth. In contrast, in the $vil$ configuration, Fig.~\ref{fig:parameter}(b), a higher number of submap factors enhances the accuracy. However, finer voxel resolutions do not always yield better trajectory estimates due to noise in the occupancy fields and gradients. Additionally, more keyframes and IMU frames do not significantly improve accuracy, as submap factors are dominant in $vil$ mode. Overall, our current parameter configuration shows a good balance between accuracy and computation time in both, depth and LiDAR, setups.

\setlength{\tabcolsep}{6.0pt}
\begin{table}
\caption{Maximum memory [GB] required for different benchmarked methods and datasets} \label{tab:mem_req}
\centering
\renewcommand{\arraystretch}{1.3} 
\begin{tabular}{c|c c c c}
     & \rotatebox[origin=c]{90}{\textit{Ours-vi}} & \rotatebox[origin=c]{90}{\textit{Ours-vid}}& \rotatebox[origin=c]{90}{\textit{Ours-vil}} & \rotatebox[origin=c]{90}{\textit{FAST-LIVO}} \\

    \hline
    \texttt{HILTI-Oxford: exp06} & 4.86 & 11.62 & 11.51 & 3.97  \\
    \texttt{VBR: Campus1} & 4.50 & 10.8 & 25.5 & 15.0  \\

    \hline
\end{tabular}
\end{table}

\subsection{Timing and memory}

In this section we demonstrate that our system can run in realtime and that the memory efficiency of the overall architecture and its modules make this system suitable to perform large-scale SLAM. We first present in Fig.~\ref{fig:timings} the running times of the main components of our SLAM system: visual frontend, realtime graph optimization, submapping interface, and the depth network. All timings were measured in a Desktop with i7-13700 CPU and RTX-3080 $10\,\text{GB}$ GPU. Thanks to our multi-threaded approach and by leveraging internal queues, we computationally decouple our submapping interface from the state estimator. Therefore, not all running times are sequential and a direct summation of these does not correspond to the overall time needed to process a multi-frame. 

In \texttt{MH05} of EuRoC, our visual-inertial configuration, \textit{Ours-vi}, runs up to $47\,\text{Hz}$, and with the depth network configuration, the MVS thread can process 8-view images at $13\,\text{Hz}$. Note that the stereo network and depth fusion with the MVS network run on a GPU. Furthermore, the networks and map integration do not necessarily have to process every frame. In \texttt{exp06} of the Hilti-Oxford dataset, \textit{Ours-vi} can process all five incoming images at $10\,\text{Hz}$, and \textit{Ours-vid} processes at $8\,\text{Hz}$ given that the bottleneck is the realtime optimization graph. The realtime estimator of \textit{Ours-vil} runs at $8\,\text{Hz}$, while the submapping interface runs at $3.5\,\text{Hz}$ due to a high volume of LiDAR points. In \texttt{Campus1}, \textit{Ours-vi} runs at $16\,\text{Hz}$, and the global position fusion from GNSS only adds negligible overhead. Note the use of high-resolution cameras in VBR leads to higher numbers of extracted keypoints. As mentioned in Sec.~\ref{sec:vbr}, we adopted additional refinement steps~\cite{xu2023unifying} in the stereo network which increases the inference time. Overall, our uncertainty-aware depth networks are capable of realtime processing at a minimum of $13\,\mathrm{Hz}$ across all datasets. However, we emphasize that many parameters offer a trade-off to exchange a little accuracy for faster runtime, which allows us to run the system including networks, depth or LiDAR integration also on a drone featuring an NVIDIA Orin NX (see supplementary video). 

To benchmark the timings, we measured total elapsed time (wall time) and divided it by the total number of processed frames for both \textit{Ours-vi} and ORB-SLAM3. Please note that due to the different algorithmic structures of OKVIS2-X and ORB-SLAM3, a direct timing comparison of individual modules is not possible, therefore we regard a wall time comparison as the fairest metric. Our method outperforms in efficiency on \texttt{MH05} ($38.1\,\textrm{vs}.\, 64.7\,\textrm{ms}$) and performs on par on \texttt{exp06} ($112.3\,\textrm{vs}.\, 116.1\,\textrm{ms}$). In \texttt{Campus1}, although our method shows a higher runtime ($107.3\,\textrm{vs}.\, 69.1\,\textrm{ms}$), it achieves significantly better accuracy ($2.499\,\textrm{m}$) compared to ORB-SLAM3 ($7.289\,\textrm{m}$), making the trade-off worthwhile.

In Table \ref{tab:mem_req}, we present the maximum memory required of OKVIS2-X with different sensor configurations and in different datasets. As expected, leveraging our dense volumetric map increases the memory requirements over landmark-based VI SLAM. When compared with FAST-LIVO~\cite{zheng2022fast}, our approach demands more memory, since our maps store free and occupied regions volumetrically, while FAST-LIVO stores downsampled point clouds only. However, our map representation can be directly used for downstream tasks, notably safe autonomous navigation of guaranteed free-space. Nonetheless, the memory requirements of the proposed approach can be fulfilled with consumer grade computers, even on-board our drone.  In \texttt{exp06}, \textit{Ours-vid} and \textit{Ours-vil} require a similar memory usage since \textit{Ours-vid} generates more submaps, due to our overlap strategy, even though the volume mapped is larger with the LiDAR sensor. In terms of GPU memory usages, \textit{Ours-vid} only takes $3.51\,\text{GB}$ for the MVS and stereo networks.

%% file: chapters/07_conclusion.tex
\section{Conclusion}
In this paper, we have presented a multi-sensor SLAM system, OKVIS2-X, a substantial extension to the sparse landmark-based OKVIS2~\cite{OKVIS2}. To achieve highly robust and accurate localization and mapping in large-scale environments, we introduced volumetric occupancy submapping that is tightly-coupled with the state estimator. Furthermore, we presented a unified framework to embrace multi-modality including visual, inertial, depth from neural networks, point clouds from a LiDAR, and global position measurements from a GNSS receiver. We have thoroughly evaluated OKVIS2-X in multiple benchmark datasets in terms of trajectory and mapping accuracy, memory usage, and timings where we have created a new baseline over the state-of-the-art SLAM systems. Notably, we could map dense volumetric occupancy while having sub-meter trajectory accuracy in a $9 \, \text{km}$ driving sequence with a visual-inertial-LiDAR configuration.

In the future, we would like to push boundaries further by incorporating online calibration of sensor time offsets, LiDAR and GNSS antenna extrinsics, multi-session relocalization, open-vocabulary scene understanding, dynamic objects, and change detection over time in complex and large-scale environments. Furthermore, adopting an on-demand local online map saving and loading scheme would offer the potential for even finer occupancy resolutions and practically unlimited map scalability. 

%% file: chapters/99_appdx.tex
\appendix

\subsection{Error State Definition}
We generally use the perturbation $\delta \mbf{\chi}_T = [\delta \mbf{r}, \delta \mbf{\alpha}]$ for poses around linearisation points for translation $\mbfbar{r}$ and orientation $\mbfbar{C}$:
\begin{equation*}
\label{eq:error-states}
    \begin{aligned}
    \mbf{r} &= \bar{\mbf{r}} + \delta \mbf{r}, \quad &
    \mbf{C} &= \Exp {\delta \boldsymbol{\alpha}} \mbfbar{C},
    \end{aligned}
\end{equation*} 
where an additional subscript is used to indicate the respective transformation. With this, we further define the error state as
$\delta \boldsymbol{\chi} = [\delta \mbf{\chi}_{T_{WS}}, \delta\mbf{\chi}_\mathrm{sb}]$, with $\delta\mbf{\chi}_\mathrm{sb}$ denoting an additive perturbation of the speed and biases.

\subsection{GNSS Covariances}
\label{appdx:gnss-covariance}
The overall covariance matrix $\boldsymbol{\Sigma}_\mathrm{g}\attime{j}$ for the GNSS residual in Eqn.~\eqref{eq:gps res} is composed of the GNSS measurement covariance matrix $\boldsymbol{\Sigma}_\mathrm{g_0}\attime{j}$ and IMU pre-integration covariance denoted by $\boldsymbol{\Sigma}_{x}\attime{j}$. 
Thus, $\boldsymbol{\Sigma}_\mathrm{g}\attime{j}$ can be derived by: 
\begin{equation*}
\label{eq:overall-gps-covariance}
    \boldsymbol{\Sigma}_\mathrm{g}\attime{j} = \boldsymbol{\Sigma}_\mathrm{g_0}\attime{j} + \mathbf{J} \boldsymbol{\Sigma}_{\mathbf{x}}\attime{j} \mathbf{J}^{T}.
\end{equation*}
Here, $\mathbf{J}$ denotes the Jacobian of  $\mathbf{e}_\mathrm{g}^{j} $ with respect to $\hat{\boldsymbol{\chi}}_{{T}_{WS_j}}$.

\subsection{GNSS Jacobians}
\label{appdx:gnss-jacobians}
The Jacobians of the global position residuals in Eqn.~\eqref{eq:gps res} with respect to the pose error states of the predicted pose $\hat{\boldsymbol{T}}_{WS}$ based on the IMU measurements and the relative pose between the GNSS and world frame $\boldsymbol{T}_{GW}$ are given by:
\begin{equation*}
\label{eq:error-jacobians}
    \begin{aligned}
        \frac{ \partial \mathbf{e}_\mathrm{g}\attime{j} }
         { \partial \delta \hat{\boldsymbol{\chi}}_{{T}_{WS_j}}}
        &= 
        \begin{bmatrix}
            -\Cbar{G}{W}
            &
            \Cbar{G}{W} \crossmx{ \Cbar{W}{S} \pos{S}{A} }
        \end{bmatrix},
        \\
        \frac{ \partial \mathbf{e}_\mathrm{g}\attime{j} }
         { \partial \delta \boldsymbol{\chi}_{T_{GW}}}
        &= 
        \begin{bmatrix}
            -\mathbf{I}_{3\times3}
            &
            \crossmx{ \Cbar{G}{W} \left( \posbar{W}{S} + \Cbar{W}{S}\pos{S}{A}\right) }
        \end{bmatrix}.
    \end{aligned}
\end{equation*}
The predicted poses are not estimated states in the posegraph optimization. Since the predicted measurement is a function of the estimated states at time step $i$, i.e.\ $\mbfhat{x}^j= f(\mbf{x}^i)$, we can actually optimize $\mbf{x}^i$ in the estimated states. The Jacobian with respect to $\mbf{x}^i$ can be computed using the chain rule and standard IMU pre-integration Jacobians, see\mbox{\cite{forster2016manifold}}.

\subsection{Submap Alignment Jacobian}
\label{appdx:submap-alignment-jacobian}
With the dense alignment residuals (Eqn.~\eqref{eq:method_lidar_residual}) only depending on pose states, but not on speed and biases, we can compute Jacobians with respect to the poses of frames $\cframe{S_{a}}$ and $\cframe{S_{b}}$ leveraging the chain rule:
\begin{equation*}
    \label{eq:jacobian_chain_rule}
    \frac{\partial e^{a,b}_{\mathrm{m}}}{\delta \mbf{\chi}_{T_{WS_k}}} 
    = 
    \frac{\partial e^{a,b}_{\mathrm{m}}}{\partial \mvec{S_a}{p}}
    \frac{\partial \mvec{S_a}{p}}{\delta \mbf{\chi}_{T_{WS_k}}},
\end{equation*}
with $k \in \{ a, b \}$. The individual Jacobians are:
\begin{align*}
    \label{eq:jacobians}
    \frac{\partial e^{a,b}_{\mathrm{m}}}{\partial \mvec{S_a}{p}} &= 
    \frac{\nabla L}{\sqrt{\frac{L_{\mathrm{min}}^{2}}{9} + \left| \nabla L \right|^{2}\sigma_{d}^{2}}}
    \nonumber \\
    \frac{\partial \mvec{S_a}{p}}{\delta \mbf{\chi}_{T_{WS_a}}} &= \Cbar{S_a}{W}
    \begin{bmatrix}
      -\mbf{I}_3 & \crossmx{{\Cbar{W}{S_b}}^\mvec{S_b}{p} + \posbar{W}{S_b} - \posbar{W}{S_a}}
    \end{bmatrix}
    \nonumber \\
    \frac{\partial \mvec{S_a}{p}}{\delta \mbf{\chi}_{T_{WS_b}}} &= \Cbar{S_a}{W}
    \begin{bmatrix}
      \mbf{I}_3 & - \crossmx{{\Cbar{W}{S_b}}\mvec{S_b}{p}}
    \end{bmatrix}
    .
\end{align*}

%% file: chapters/999_supplementary.tex
\clearpage

\section*{Supplementary Material}
In the supplementary material, we provide additional details that were omitted from the main paper due to space limitations. These primarily concern the ablation study on online extrinsic calibration (Sec.~\ref{sec:abl_calib}) and the demonstration of GNSS fusion using real-world GNSS data (Sec.~\ref{sec:eval-gnss}).

\begin{figure*}[b]
\label{fig:ext-details}
\centering
\subfloat[ ]{\includegraphics[width=.33\linewidth]{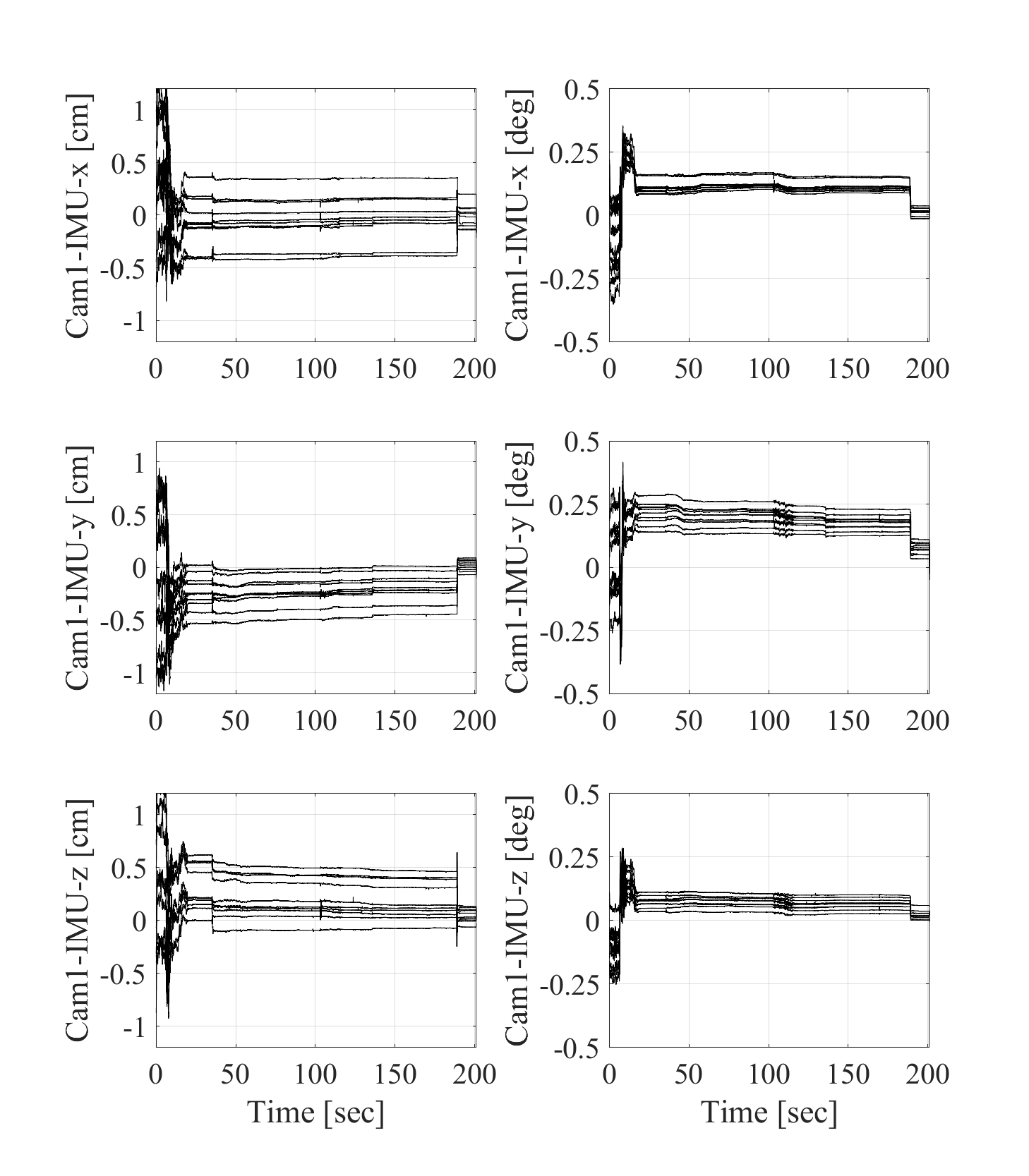}%
\label{fig:cam1-ext}}
\hfil
\subfloat[ ]{\includegraphics[width=.33\linewidth]{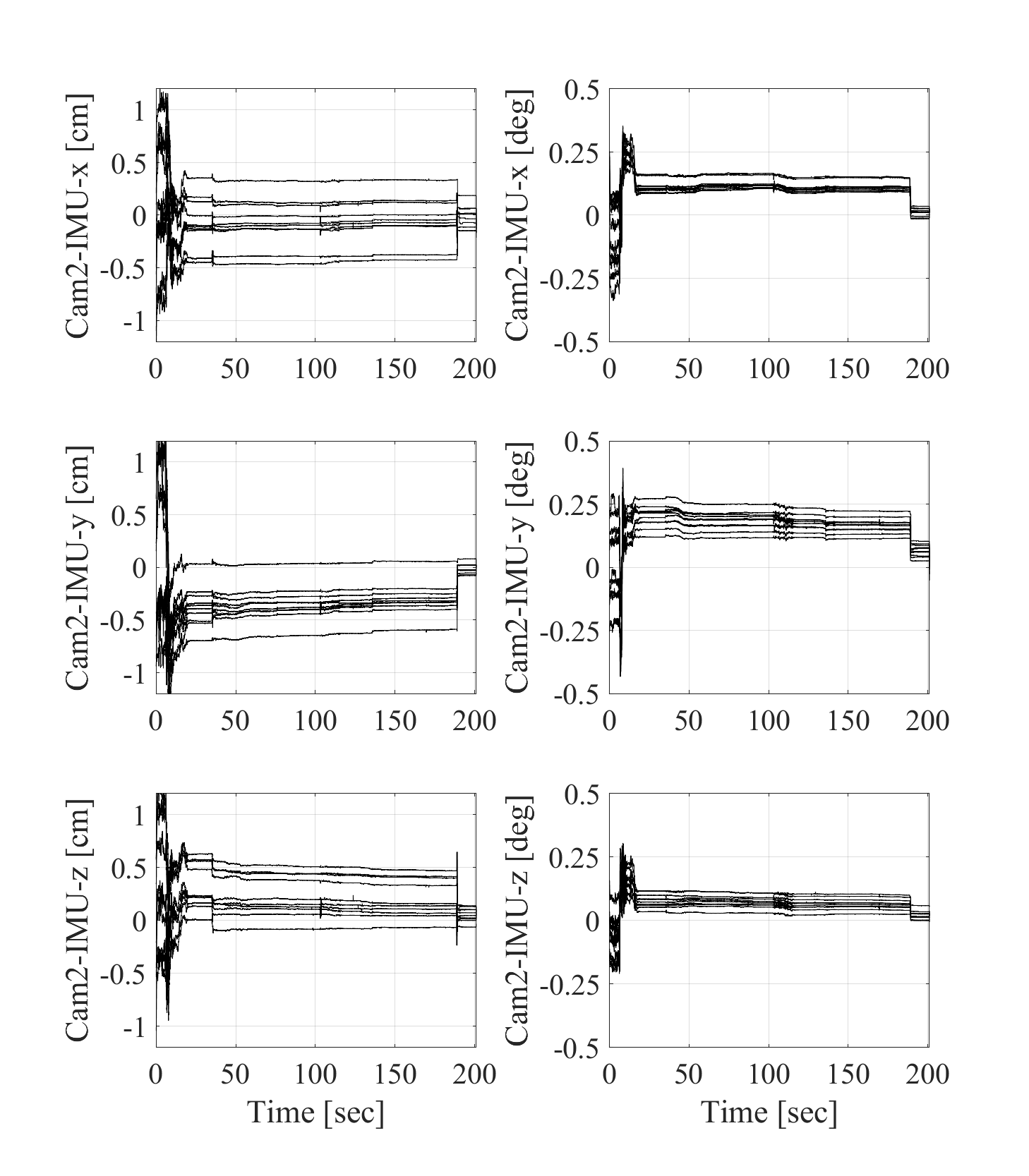}%
\label{fig:cam2-ext}}
\hfil
\subfloat[ ]{\includegraphics[width=.33\linewidth]{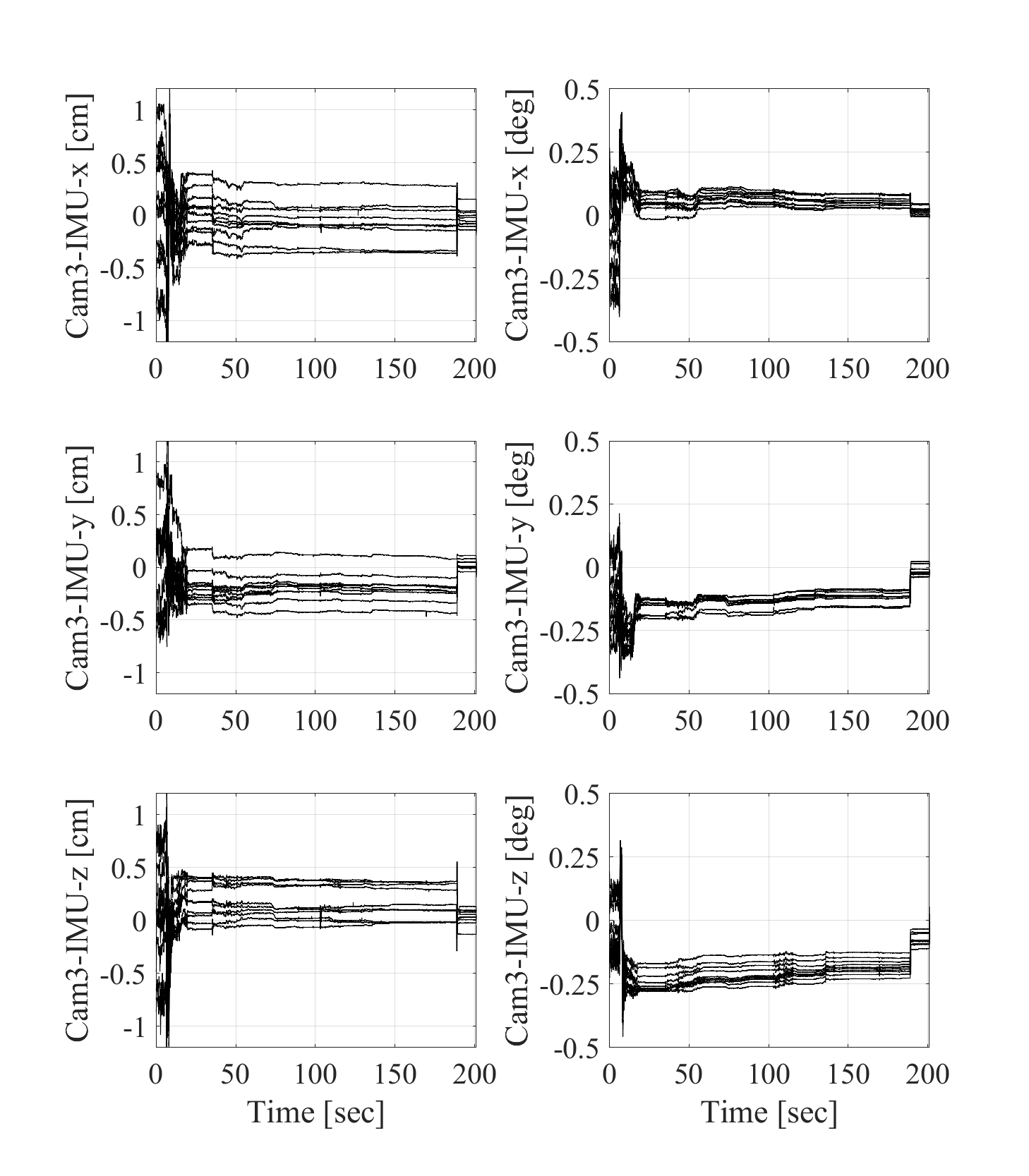}%
\label{fig:cam3-ext}}

\vfil
\subfloat[ ]{\includegraphics[width=.33\linewidth]{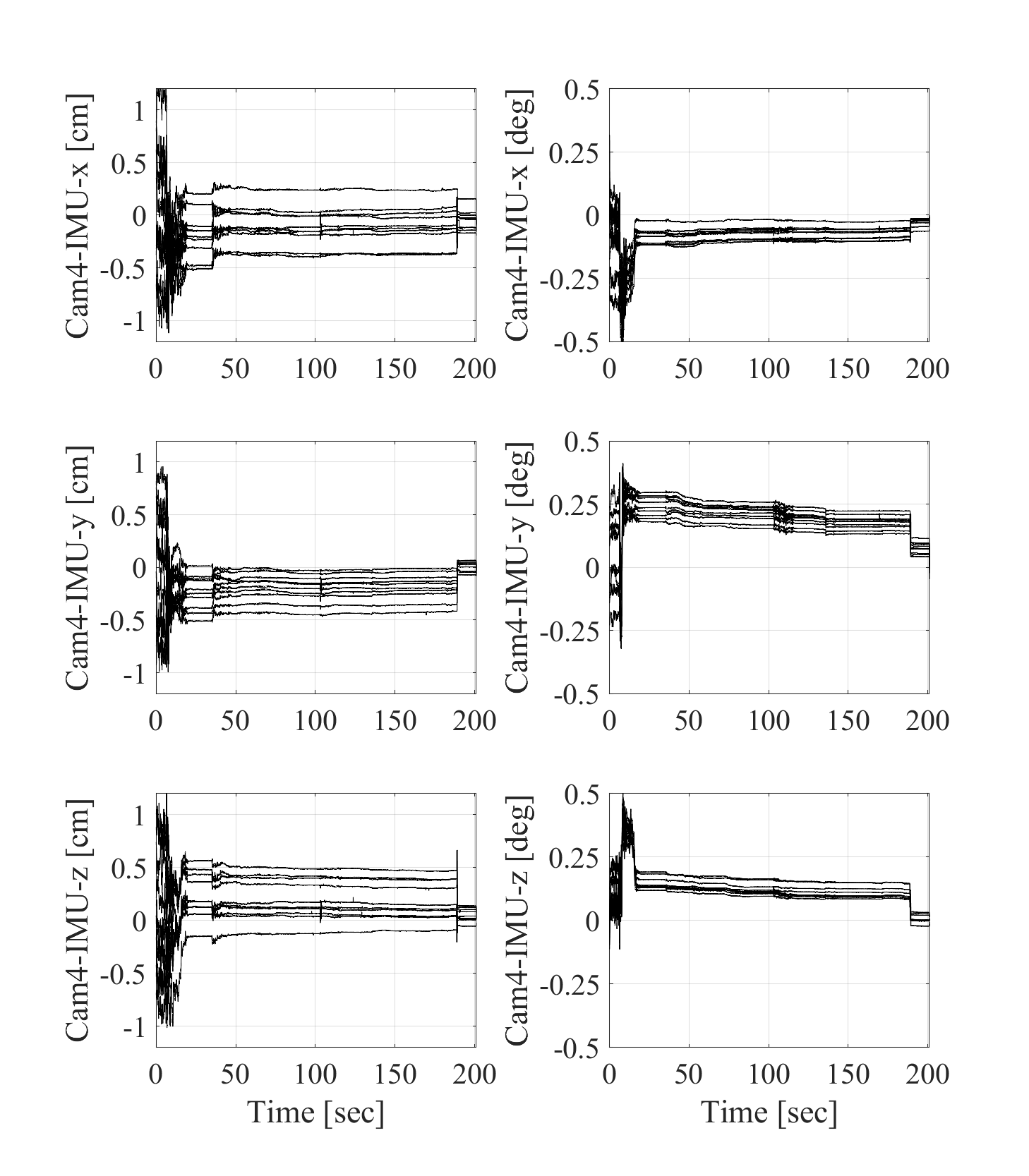}%
\label{fig:cam4-ext}}
\hfil
\subfloat[ ]{\includegraphics[width=.33\linewidth]{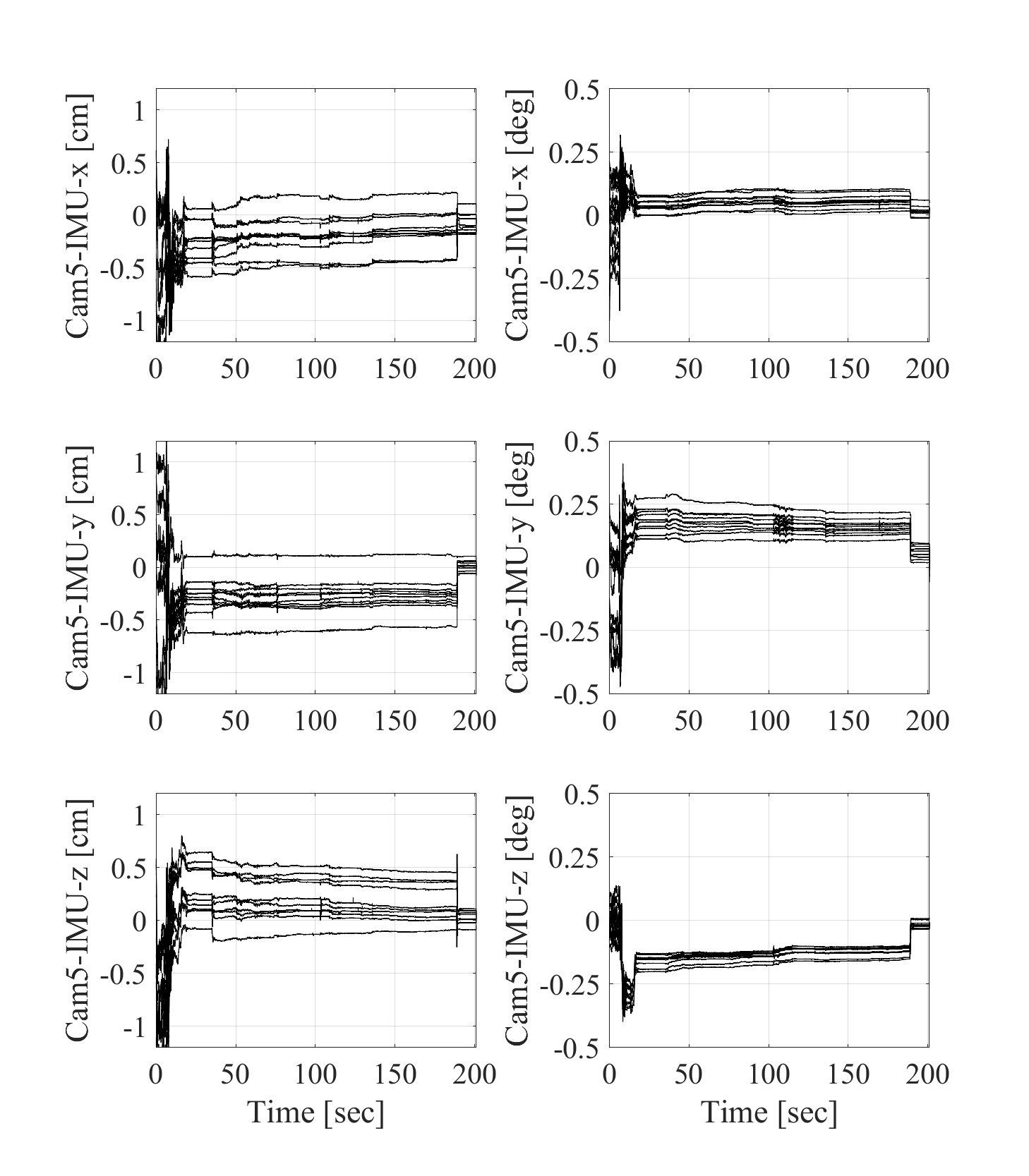}%
\label{fig:cam5-ext}}
\caption{Camera-IMU extrinsic perturbation study in \texttt{exp06}, Hilti-Oxford dataset for 5 cameras. Each line represents the difference between online calibration and the post-calibrated extrinsics in 10 independent runs.}
\end{figure*}

\subsection*{Online extrinsic calibration - Details}
Robustness to the initial extrinsic error is crucial in practice.
To show the robustness of OKVIS2-X to the initial perturbation of the extrinsic prior, we conducted an additional experiment which was only briefly summarized at the end of Sec.~\ref{sec:abl_calib} and is detailed further here.

The contribution in multi-camera to IMU extrinsic calibrations is shown on sequence \texttt{exp06} in the Hilti-Oxford dataset which contains 5 cameras. Each axis of 5 cameras $(i=1,2,\cdots,5)$ was perturbed with a uniform distribution,

\begin{align*}
    \Tbar{S}{C_i} &= \T{S}{C_i}
    \begin{bmatrix}
        \mathrm{Exp}(\delta \alpha_i) & \delta r_i
        \\ 0_{1 \times 3} & 1
    \end{bmatrix}, \\
    \mathrm{where} \:\:\: &\delta \alpha_i \sim \mathcal{U}(-0.3^\circ, 0.3^\circ), \nonumber \\
    &\delta r_i \sim \mathcal{U}(-1\,\mathrm{cm}, 1\,\mathrm{cm}). \nonumber
\end{align*}
We use the same notations as defined in the main paper. For example, $\cframe{S}$ and $\cframe{C_i}$ are IMU and $i^\mathrm{th}$ camera coordinate frames, respectively. Since we do not know the true extrinsics, the perturbed extrinsics $\Tbar{S}{C_i}$ is perturbed from our post-calibrated extrinsics $\T{S}{C_i}$ which was obtained by full bundle adjustment with all available measurements. Figures~\ref{fig:cam1-ext}-\ref{fig:cam5-ext} visualize the convergence behavior of all five camera-to-IMU extrinsics in 10 runs. The sudden correction observed after roughly 190 seconds was due to a loop closure that as such provides additional information leading to a sudden change in extrinsics. Considering 10 runs, online estimated extrinsics in OKVIS2-X show a standard deviation of $0.08\,\mathrm{cm}$ and $0.02^\circ$ at the end of the sequence. This study demonstrates the robustness of online calibration to the initial error. 

\begin{table}[b]
\centering
    \begin{threeparttable}
    \caption{RMSE ATE [m] of the different approaches in the \texttt{complex\_environment} sequence.}
    \begin{tabular}{|c|c|c|c|}
        \hline
         & Causal & Non-Causal & Final BA\\
        \hline
        RTKLIB (SPP) & $19.99$\tnote{1} & - & - \\
        \hline
        GVINS  & $4.45$ & - & - \\
        \hline
        Ours-vi & $23.72$ & $23.72$ & $23.74$ \\
        \hline
        Ours-vig & $12.51$ & $8.47$ & $5.12$ \\
        \hline
    \end{tabular}
    \begin{tablenotes}
       \item [1] RMSE ATE is evaluated for the part of the sequence that is not indoors.
    \end{tablenotes}
    \label{tab:gvins-results}
    \end{threeparttable}
\end{table}

\subsection*{GNSS Fusion - Details on Experiment}
As a first note, we want to elaborate on the definition of \textit{tightly-coupled} fusion of GNSS measurements used in this paper. We found that in the literature the definition of \textit{tightly-coupled} fusion is not 100\% clear. In our definition, we follow works like~\cite{cioffi2020tightly}, in which the term \textit{tightly-coupled} refers to a tight coupling of heterogeneous sensor modalities, such as IMU and vision, on the level of the non-linear optimization. This means, that residuals from different sensor sources are all minimized in one unified optimization problem as opposed to e.g.,\ ~\cite{VINS-Fusion}, where global fusion is performed as a separate, loosely-coupled, parallel posegraph optimization.

Furthermore, we want to give some more insights into the experiment in Sec.~\ref{sec:eval-gnss}, where the capability of OKVIS2-X to handle real-world, and potentially erroneous, GNSS data has been demonstrated. The experiment was conducted on the dataset published by the authors of GVINS~\cite{GVINS} alongside their original paper (\url{https://github.com/HKUST-Aerial-Robotics/GVINS-Dataset}). This dataset consists of a stereo camera, an IMU and a ZED-F9P GNSS sensor. The authors provide RTK solutions as well as raw GNSS measurements (e.g.\ pseudoranges, Doppler measurements, etc.). 
As stated in our paper, we chose to evaluate on the nearly $3$ km \texttt{complex\_environment} sequence due to its numerous challenges for different sensor modalities: featureless images in very bright or dark areas as well as high corruption or complete unavailability of GNSS signals in cluttered environments or indoors (see Fig.~{\ref{fig:complex-gvins}). Using RTKLIB~\cite{rtklib}, we computed the Single Point Positioning (SPP) solution as a measurement that would be available from a low-grade consumer-level GNSS device, and used it as an input for the proposed GNSS fusion approach. It is worth mentioning that we could not achieve the same level of SPP accuracy as originally stated in the paper. The RMSE ATE of the SPP solution with respect to the RTK solution in areas with a valid RTK fix is $19.99$m as opposed to $6.036$m as stated in ~\cite{GVINS}. Even after filtering out outliers with an error of more than $40$ meters, the RMSE ATE of the SPP solution would still be $9.01$m. Table~\ref{tab:gvins-results} summarizes the resulting accuracy of OKVIS2-X (using the unfiltered SPP solutions), GVINS and SPP. The RMSE ATE has been computed after a full 6DoF alignment of the trajectories. Note that for a fairer comparison with GVINS, OKVIS2-X has also been evaluated without visual loop closures.
This evaluation shows that, even though the system was originally designed to fuse intermittent high-accuracy (RTK-) GNSS measurements, it can still also benefit from lower-grade GNSS measurements.

However, we want to emphasize that the proposed system does not claim superiority for the fusion of GNSS in general. We recognize the impressive work of GVINS and think that adopting their formulations for error terms based on the raw measurements, such as pseudoranges and Doppler measurements could improve our system significantly.